\documentclass[10pt,a4paper]{amsart}

\usepackage{anysize}
\marginsize{20mm}{20mm}{20mm}{15mm}
\usepackage[utf8]{inputenc}
\usepackage[T1]{fontenc}
\usepackage{amsmath,amsfonts,mathrsfs,bm}

\usepackage[table]{xcolor}% http://ctan.org/pkg/xcolor

\usepackage[pdftex]{graphicx}
\usepackage[title]{appendix}
\usepackage{amsthm}
\usepackage[vlined,boxruled]{algorithm2e}
\usepackage{enumitem}
\usepackage{blindtext}
\usepackage{mdframed}
\usepackage[sort,square,numbers]{natbib}
\usepackage{multicol}

\usepackage{dsfont}

\usepackage{hyperref}
\usepackage[capitalise]{cleveref}

\usepackage[labelformat=simple]{subcaption}
\usepackage[justification=centering]{caption}

\usepackage{booktabs}

\usepackage{tikz}
\usetikzlibrary{arrows.meta}
\usetikzlibrary{positioning}
\usetikzlibrary{shapes,calc}
\usetikzlibrary{arrows,fit}
\usetikzlibrary{shapes.multipart}

\usepackage{adjustbox}

\usepackage{array}
\newcolumntype{L}[1]{>{\raggedright\let\newline\\\arraybackslash\hspace{0pt}}m{#1}}
\newcolumntype{C}[1]{>{\centering\let\newline\\\arraybackslash\hspace{0pt}}m{#1}}
\newcolumntype{R}[1]{>{\raggedleft\let\newline\\\arraybackslash\hspace{0pt}}m{#1}}

\usepackage{multirow,multicol}

\usepackage{float}
\usepackage{pgffor}

\usepackage[section]{placeins}

\usepackage{chngcntr}
\counterwithin{figure}{section}
\counterwithin{table}{section}

\newmdenv{allfour}
\newmdenv[leftline=false,rightline=false]{topbot}
\newmdenv[topline=false,rightline=false]{leftbot}
\newmdenv[topline=false,leftline=false]{rightbot}
\newmdenv[topline=false,rightline=false,leftline=false]{bottom}

\newtheorem{theorem}{Theorem}[section]

\theoremstyle{definition}

\theoremstyle{remark}
\newtheorem{remark}[theorem]{Remark}

\providecommand{\keywords}[1]{\textbf{\textit{Keywords -- }} #1}

\newcounter{defcounter}
\definecolor{obscol}{HTML}{288A42}
\definecolor{unobscol}{HTML}{000000}
\definecolor{lstm}{HTML}{B83232}
\definecolor{srnn}{HTML}{28568A}
\definecolor{trm}{HTML}{288A42}
\definecolor{data}{HTML}{000000}
\definecolor{ecol}{HTML}{000000}
\definecolor{pcol}{HTML}{000000}

\newsavebox{\obs}
\savebox{\obs}{%
	\begin{tikzpicture}[
	>=stealth,
	]
	\draw[thick, black] (0,0) --(0,1);
	\draw [fill=obscol] (0,0) circle (1.5pt);
	\draw [fill=obscol] (0,0.167) circle (1.5pt);
	\draw [fill=unobscol] (0,0.333) circle (1.5pt);
	\draw [fill=unobscol] (0,0.5) circle (1.5pt);
	\draw [fill=obscol] (0,0.667) circle (1.5pt);
	\draw [fill=unobscol] (0,0.833) circle (1.5pt);
	\draw [fill=obscol] (0,1) circle (1.5pt);
	\end{tikzpicture}%
}

\newsavebox{\all}
\savebox{\all}{%
	\begin{tikzpicture}[
	>=stealth,
	]
	\draw[thick, black] (0,0) --(0,1);
	\draw [fill=obscol] (0,0) circle (1.5pt);
	\draw [fill=obscol] (0,0.167) circle (1.5pt);
	\draw [fill=obscol] (0,0.333) circle (1.5pt);
	\draw [fill=obscol] (0,0.5) circle (1.5pt);
	\draw [fill=obscol] (0,0.667) circle (1.5pt);
	\draw [fill=obscol] (0,0.833) circle (1.5pt);
	\draw [fill=obscol] (0,1) circle (1.5pt);
	\end{tikzpicture}%
}

\newcommand{\R}{\mathbb{R}}
\newcommand{\peq}{.}
\newcommand{\veq}{,}

\newcommand{\q}[1]{``#1''}

\newcommand{\di}{\,\mathrm{d}}

\newcommand{\Nin}{N_{\textrm{in}}}
\newcommand{\Nout}{N_{\textrm{out}}}
\newcommand{\Nex}{N_{\textrm{ex}}}
\newcommand{\Nseq}{N_{\textrm{seq}}}
\newcommand{\Nst}{N_{\textrm{state}}}

\renewcommand{\arraystretch}{1.75}

\SetAlgoSkip{bigskip}

\newcommand{\nps}{10}
\newcommand{\psa}{3}
\newcommand{\psb}{5}
\newcommand{\psc}{10}

\definecolor{myred}{HTML}{B83232}
\definecolor{myblue}{HTML}{28568A}
\definecolor{mygreen}{HTML}{288A42}

\begin{document}
	
	%% Title
	\title{Short-term traffic prediction using physics-aware neural networks}
	
	\author[M.~Pereira]{Mike Pereira} \address[Mike Pereira]{\newline Department of Mathematical Sciences \& Department of Electrical Engineering
		\newline Chalmers University of Technology \& University of Gothenburg
		\newline S--412 96 G\"oteborg, Sweden.} \email[]{mike.pereira@chalmers.se}
	
		\author[A.~Lang]{Annika Lang} \address[Annika Lang]{\newline Department of Mathematical Sciences
		\newline Chalmers University of Technology \& University of Gothenburg
		\newline S--412 96 G\"oteborg, Sweden.} \email[]{annika.lang@chalmers.se}
	
	\author[B.~Kulcs\'ar]{Bal\'azs Kulcs\'ar} \address[Bal\'azs Kulcs\'ar]{\newline Department of Electrical Engineering
		\newline Chalmers University of Technology
		\newline S--412 96 G\"oteborg, Sweden.} \email[]{kulcsar@chalmers.se}

	\thanks{
		Acknowledgement.  The authors thank G.~Szederkenyi and Gy.~Liptak for the helpful discussions about the TRM, as well as P.~Boyraz Baykas for her inspiring inputs. 
		The authors thank for their financial support the Chalmers AI Research Centre (CHAIR) (under the projects STONE, RITE, and SCNN) and the Transport Area of Advance of the Chalmers University of Technology (under the projects STONE and IRIS). This work has been partially supported and funded by OPNET (Swedish Energy Agency, 46365-1), the Wallenberg AI, Autonomous Systems and Software Program (WASP) funded by the Knut and Alice Wallenberg Foundation, and the Swedish Research Council (project nr.\ 2020-04170). The traffic data are provided by, and used with the permission of Rijkswaterstaat -- Centre for Transport and Navigation (Netherlands).
	}
	
	\subjclass{}
	\keywords{}

	\maketitle
	
	\begin{abstract}
		In this work, we propose an algorithm performing short-term predictions of the flux of vehicles on a stretch of road, using past measurements of the flux. This algorithm is based on a  physics-aware recurrent neural network. A discretization of a macroscopic traffic flow model (using the so-called Traffic Reaction Model) is embedded in the architecture of the network and yields  flux predictions based on estimated and predicted space-time dependent traffic parameters. These parameters are themselves obtained using a succession of LSTM ans simple recurrent neural networks. Besides, on top of the predictions, the algorithm  yields a smoothing of its inputs which is also physically-constrained by the macroscopic traffic flow model. The algorithm is tested on raw flux measurements obtained from loop detectors. 
	\end{abstract}
	
	%-----------------------------------------------------------------------------------------------------------
	%-----------------------------------------------------------------------------------------------------------
	
	\section{Introduction}
	
	Using neural networks to produce short-term traffic predictions is a subject of growing interest, especially since advances in transportation systems increased substantially the availability of large amount of traffic data from various sources \cite{zhu2018big,zhang2011data}.  We refer the interested reader to \cite{lee2021short,do2019survey} for surveys of neural networks-based models that have been proposed for short-term traffic predictions. These methods have proven to be particularly fit to tackle the challenges associated with traffic prediction such as the modeling of spatio-temporal dependencies in traffic data or the influence of external factors (e.g.\ weather and road conditions, unpredictable traffic events) \cite{vlahogianni2014short}.	
	
	Particular architectures have been proposed to model the temporal dependencies present in traffic data (see \cite{yin2020comprehensive} for a comprehensive survey), which are due for instance to the fact that these data often consist of time series of measurements of traffic-related quantities (such as the occupancy, speed and flux)  made by sensors on a road. Consequently, algorithms taking sequential data as input have been proposed to reflect for traffic predictions: the input then consists of a sequence of (possibly multivariate) variables corresponding to measurements at some times $t_{-(N_p-1)} \le \dots \le t_{-1} \le t_0$ and generates predictions of some quantity of interest at times $t_1 \le \dots \le t_{N_f}$ (where $t_1 \ge t_0$). Examples of such algorithms include time-delayed neural networks (e.g.\ \cite{zhong2005short,7569236,lingras2001time}), (graph) convolutional neural networks (e.g.\ \cite{pan2019urban,yu2017spatio}) and recurrent neural networks (RNNs) (e.g.\ \cite{xu2017real,ma2015long,wondimagegnehu2017short}), which will be of particular interest in this work.
	
	However, regardless of their great prediction performances, a caveat of these approaches is their black-box nature: indeed, due to their complex nature and high number of parameters, these models lack interpretability as it is hard to understand how (and if) these algorithms manage to produce reliable traffic predictions \cite{vlahogianni2014short,do2019survey}.  Also, these algorithms (and their performance) depend on a number of hyperparameters (such as the number of hidden layers, the number of nodes,...) that must be tuned by the practitioner in order to achieve the best performances for a given dataset. 
	
	Bearing this in mind, we propose in this work a \q{grey-box} RNN architecture which aims at providing short-term flux predictions based on past flux measurements along a stretch of road. The originality of this architecture is that it combines a conventional RNN with a discretization of a macroscopic traffic flow model in an attempt to yield interpretable predictions. Indeed, the space-time varying parameters of the traffic model are seen as a set of unobserved time series which are learned from the flux observations. In a first step, a conventional RNN is used to extract these time series of parameters from the observations and predict their evolution over a short period of time. The output of this first algorithm is then passed to the traffic model to yield both a re-estimation of the flux observations and their flux predictions. Besides, since the macroscopic model is defined on the whole road, flux estimates and predictions can be obtained everywhere on the road, hence providing the practitioner with an algorithm able to complete and predict the flux variations along a road.
	
	Note that the traffic predictions obtained with the proposed approach come directly from the (discretized) macroscopic traffic model, and therefore are bound to follow the associated physical model. The idea of constraining neural networks with physical models is receiving much attention in the context of solving forward and inverse problems for partial differential equations (PDEs) \cite{pakravan2021solving}. Two main approaches have been proposed in this context: 
	\begin{itemize}
		\item either augmenting the cost function used to train the neural network with terms that describe the PDE and its possible boundary or initial conditions (e.g.\ \cite{raissi2017physics,xu2019neural}),
		\item or explicitly embedding the PDE inside the architecture of the neural networks by dedicating some parts of the network to the approximation of differential operators (e.g.\ \cite{long2018pde,long2019pde}) or to the numerical approximation of the PDE itself (e.g.\ \cite{berg2017neural,dal2020data}).
	\end{itemize}
	In particular, Pakravan et al. \cite{pakravan2021solving} proposed so-called physics-aware neural networks for solving inverse PDE problems, which, much like our approach, consist of using a neural network to learn the parameters of a PDE, and then inputting these parameters into a numerical scheme designed to approximate the PDE. Hence, our approach can be seen as an extension of their approach for prediction purposes and in the presence of real (traffic) data.

	Another original aspect of this research is the use of a particular discretization scheme for the macroscopic traffic model: the so-called Traffic Reaction Model \cite{liptak2020traffic}. This scheme models flow dynamics along a discretized road as a chain of chemical reactions happening between adjacent cells (now seen as compartments containing \q{fictional} chemical reactants).  This scheme was shown to be able to reproduce complex traffic pattern observed when performing state and parameter estimation from real-world traffic data \cite{Pereira2021ParameterAD}. It also allows to assimilate the space-time varying parameters of the macroscopic traffic model to dimensionless and bounded parameters that are interpreted as the reaction rates of the chemical reactions mentioned above.

	In addition to the interpretation that can be given to the parameters of our architecture and the origin of the predictions it produces, most of its hyperparameters are set by the configuration of the detectors along the road of interest, hence removing this burden of tuning them. Moreover, our approach is particularly suited to high resolution data (i.e.\ data composed of measurements made with high frequency, for instance every minute or less). Such datasets are known to be problematic in prediction applications, due to the high level of noise they might have \cite{LI20111006}. In such cases, noise reduction techniques, such as smoothing or wavelet algorithms (e.g.\ \cite{jiang2004wavelet}),  or aggregation of the measurements are usually applied to the data before proceeding to the predictions. But there is no consensus on which approach to select (and with which parameters), and besides, these techniques have be shown to impact and distort the properties of time series \cite{vlahogianni2011temporal}. Our approach on the other hand naturally incorporates a smoothing of the measurements inherited from the discretized macroscopic traffic model, and is therefore physically justified.
	
		In summary, we propose a physics-aware algorithm for short-term traffic predictions. This algorithm produces a smoothing of the input data and predictions (even on unobserved sections of the road) that are highly interpretable as they arise directly from a (discretization of a) macroscopic traffic flow model. Besides, both the architecture and the parameters of our algorithm find physical interpretations.
		
	The outline of this paper is as follows. In \Cref{sec:mdl}, we present the macroscopic traffic model and its discretization used in the flux prediction architecture. In \Cref{sec:NN}, we provide some reminders on RNNs and present the architecture. Finally, in \Cref{sec:num}, we present numerical experiments based on a dataset of flux measurements taken on a Dutch motorway.

	\section{Traffic models and data}
	\label{sec:mdl}
	
	\subsection{First order macroscopic traffic flow model}
	
	First order macroscopic traffic flow models link the evolution of two quantities describing the flow and state of the traffic at any point in time and space: the flux of vehicles $f$ and the density of vehicles $\rho$. Both quantities can in particular be coupled through a conservation law expressing a balance between the numbers of vehicles entering and exiting any section of the road and the number of vehicles present in this same section. Such balance laws are common when studying for instance the flow of incompressible fluids and take the following form:
	\begin{equation}
	\partial_t \rho(t,x) + \partial_xf(t,x)= q(t,x), \quad t\ge 0, \quad x\in \R \veq
	\label{eq:lwr}
	\end{equation}
	where $q$ is a function acting as a source term and describing for instance on- and off-ramps along the road.
	
	One of the most popular macroscopic traffic flow model, the so-called Lighthill--Whitham--Richards (LWR) model \citep{richards,lighthill1955kinematic}, takes the additional assumption that the flux of vehicles can be expressed as a function of the density of vehicles. A popular choice of flux function is the so-called Greenshield flux \citep{greenshields1935study}, for which $f$ takes the form
	\begin{equation}
	f(t,x)=4f_m(t,x)\times\frac{\rho(t,x)}{\rho_m}\left(1-\frac{\rho(t,x)}{\rho_m}\right), \quad t\ge 0, \quad x\in \R, \quad \rho\in [0,\rho_m] \veq
	\label{eq:greenf}
	\end{equation}
	where $\rho_m>0$ denotes the maximal value the density can take (i.e.\ the density value corresponding to a bumper-to-bumper traffic) and $f_m\ge 0$ denotes the (possibly space-time-dependent) maximal flux of vehicles. This choice follows naturally when enforcing two straightforward and physically meaningful conditions that a flux function should satisfy, namely that the flux should be $0$ when the road is empty (i.e.\ $\rho=0$) or at full capacity (i.e.\ $\rho=\rho_m$). 
	
	Note that considering a space-time-dependent maximal flux $f_m$ allows to model for instance situations where the traffic flow is impacted by road conditions or policies that may vary in  time or space. For any given bounded and integrable initial condition $\rho_0 \equiv \rho(0,\cdot)$, and any $T>0$, the existence and uniqueness of (entropy) solutions of the resulting PDE in $C([0,T), L^1(\R)\cap L^\infty(\R))$ has been proven in the general case, where the maximal speed is taken to be a function $(t,x)\mapsto f_m(t,x)$ that is bounded and Lipschitz-continuous \citep{karlsen2004convergence,chen2005quasilinear}.

	\subsection{Solving numerically the PDE: the Traffic Reaction Model}
	\label{sec:trm}
	
	In general, no closed-form solution of PDE~\eqref{eq:lwr} is available, and therefore, numerical methods must be used to approximate it. In particular, finite volume schemes have been widely used to compute solutions to (hyperbolic) PDEs  of the form~\eqref{eq:lwr} since these solutions can develop discontinuities in finite time but remain locally integrable \citep{leveque2002finite}. For reasons which will be clarified in subsequent sections of this paper, we focus on a particular scheme proposed by Liptak et al. \cite{liptak2020traffic} and  called {Traffic Reaction Model} (TRM). This scheme can be seen as a particular instance of the finite volume schemes investigated in \cite{chainais2001finite}.

	Let us once again consider PDE~\eqref{eq:lwr} on $(\mathbb{R}_+\times \mathbb{R})$ and the following space-time discretization: in time we consider equidistant time steps $t_n=n\Delta t$ (for some step size $\Delta t>0$ and $n\in\mathbb{N}$) and in space we consider equispaced cells of size $\Delta x>0$ with centroids $x_j=j\Delta x$ (for $j\in\mathbb{Z}$). We denote by  $x_{j\pm 1/2}=x_j\pm\Delta x/2$ the boundary points of the $j$-th cell. For each cell $j$ and time step $t_n$, a quantity $\rho_j^n$ is defined as follows. At $t=t_0$, 
	\begin{equation*}
	\rho_j^0= \frac{1}{\Delta x} \int_{x_{j-1/2}}^{x_{j+1/2}} \rho_0(x)\di x,  \quad j\in\mathbb{Z} \veq
	\end{equation*}
	where $\rho_0$ denotes the initial condition associated with PDE~\eqref{eq:lwr}, and is assumed to be known. Then, the TRM approximates the solution of PDE~\eqref{eq:lwr} at $(t_n, x_j)$ by the quantity $\rho_j^n$ obtained through the following recurrence relation
	\begin{equation}
	\rho_j^{n+1}=\rho_j^n + \rho_m\left(C_{j-1}^n F(\rho_{j-1}^n, \rho_{j}^n)-C_{j}^n F(\rho_{j}^n, \rho_{j+1}^n) \right) + q_j^n, \quad j\in\mathbb{Z}, \quad n \in\mathbb{N} \veq
	\label{eq:trm}
	\end{equation}
	where the quantities $C_j^n$ and $q_j^n$ are given by
	\begin{equation*}
	\begin{aligned}
	C_j^n &
	=\frac{4}{\rho_m\Delta x} \int_{t_n}^{t_{n+1}} f_m(t, x_{j+1/2}) \di t, \quad j \in\mathbb{Z}, \quad n\in\mathbb{N} \veq \\
	q_j^n &= \frac{1}{\Delta x}\int_{t_n}^{t_{n+1}}\int_{x_{j-1/2}}^{x_{j+1/2}} q(t,x)\di x\di t, \quad j \in\mathbb{Z}, \quad n\in\mathbb{N} \veq
	\end{aligned}
	\end{equation*}
	and $F$ denotes the numerical flux defined by
	\begin{equation*}
	F(r,r')=\frac{r}{\rho_m}\left(1-\frac{r'}{\rho_m}\right), \quad r,r' \in [0,\rho_m] \peq
	\end{equation*}
	Note in particular that, in order to apply this recurrence relation, the maximal flux $f_m$ and the source term $q$ (or at least their discretized counterparts $C_j^n, q_j^n$) must be known at any time and space location.
	
	\begin{remark}\label{rem:ode}
		Note that the recurrence relation~\eqref{eq:trm} actually stems from a stable forward Euler discretization of the system of ordinary differential equations (ODEs) satisfied by a set of functions $\lbrace \rho_j\rbrace_{j\in\mathbb{Z}}$ and given by
		\begin{equation*}
		\dot{\rho}_j(t) = \frac{1}{\Delta x}\left( f_m(t, x_{j-1/2})F(\rho_{j-1}, \rho_j)- f_m(t, x_{j+1/2})F(\rho_{j}, \rho_{j+1})\right) + Q_j(t), \quad t>t_0,
		\end{equation*}
		where for $j\in\mathbb{Z}$, $Q_j$ is given by
		\begin{equation*}
		Q_j(t)=\frac{1}{\Delta x}\int_{x_{j-1/2}}^{x_{j+1/2}} q(t,x)\di x, \quad t\ge 0 \peq
		\end{equation*}
	\end{remark}

	As defined, the quantity $\rho_j^n$ acts as an approximation of the cell average at time $t_n$ and on the $j$-th cell of the solution $\rho$ of PDE~\eqref{eq:lwr} which is the ratio of the integral over the $j$-th cell of the function $x\mapsto \rho(t_n, x)$ to the cell size. Assuming that $f_m$ is bounded and Lipschitz-continuous,  Chainais-Hillairet and  Champier \cite{chainais2001finite} proved the convergence of a class of finite volume schemes that includes the TRM as the step size $\Delta t, \Delta x \rightarrow 0$, provided that these quantities satisfy a so-called Courant--Friedrichs--Lewy (CFL) condition given by
	\begin{equation}
	\frac{\Delta t}{\Delta x} < \frac{\rho_m}{8 \Vert f_m \Vert_{\infty}} \veq
	\label{eq:cfl}
	\end{equation}
	where $\Vert f_m \Vert_{\infty}=\sup\limits_{t\ge 0, x\in\R} \vert f_m(t,x)\vert$. Note that since $f_m$ is nonnegative, this condition imposes that the coefficients $C_j^n$, $j\in\mathbb{Z}$, $n\in\mathbb{N}$, satisfy
	\begin{equation}
	C_j^n < \frac{1}{2} , \quad j \in\mathbb{Z}, \quad n\in\mathbb{N} \peq \label{eq:cfl_coef}
	\end{equation}
	This condition is instrumental into proving the stability, monotonicity and ultimately convergence of the TRM scheme. 
	
	\begin{remark}\label{rem:flux_approx}
		Multiplying the recurrence relation by $\Delta x$ yields an equation describing the variation of the number of vehicles inside the $j$-th cell between $t_n$ and $t_{n+1}$:
		\begin{equation*}
		\rho_j^{n+1}\Delta x-\rho_j^n\Delta x = (\rho_m\Delta x) C_{j-1}^n F(\rho_{j-1}^n, \rho_{j}^n)- (\rho_m\Delta x)C_{j}^n F(\rho_{j}^n, \rho_{j+1}^n) + q_j^n\Delta x.
		\end{equation*}
		In this last equation, the quantity $q_j^n\Delta x$ corresponds to the net volume of vehicles added or subtracted to the cell by the source term, and the quantity $(\rho_m\Delta x) C_{j}^n F(\rho_{j}^n, \rho_{j+1}^n)$ corresponds to the volume of vehicles exiting the cell $j$ and entering the cell $(j+1)$. Hence, dividing this last quantity by the time step $\Delta t$ then gives an approximation of the flux of vehicles $f(t_n, x_{j+1/2})$ at the interface between the cells $j$ and $(j+1)$, and at time $t_n$:
		\begin{equation*}
		f(t_n, x_{j+1/2}) \approx \frac{\rho_m\Delta x}{\Delta t} C_{j}^n F(\rho_{j}^n, \rho_{j+1}^n), \quad j\in\mathbb{Z}, \quad n\in\mathbb{N}.
		\end{equation*}	
	\end{remark}
	
\begin{figure}
		\centering
		\resizebox{0.8\textwidth}{!}{
			\input{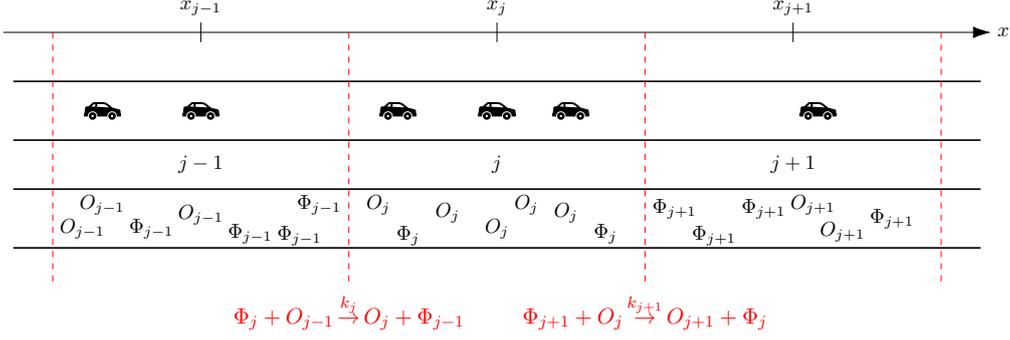}
		}
		\caption{Representation of the Traffic Reaction Model interpretation of traffic flow.}
		\label{fig:rr}
	\end{figure}

	\begin{remark}
		
		The TRM can be derived by assimilating the traffic flow dynamic (on the discretized road) to a chemical reaction network \citep{liptak2020traffic,Pereira2021ParameterAD}. 
		More precisely, each cell is modeled as a compartment containing two (homogeneously-distributed) fictional chemical reactants: molecules of occupied space $O_j$ and molecules of free space~$F_j$. The density of vehicles $\rho_j$ within a cell $j$ is then assimilated to the concentration of occupied space~$O_j$, and the variations in time of this density is interpreted as resulting from chemical reactions happening at the boundaries of the cell. Namely, the chemical reaction happening at the boundary between the cells $j$ and $j+1$ \q{transforms} a molecule of occupied space $O_j$ in $j$ and a molecule of free space $\Phi_{j+1}$ in $j+1$ into  a molecule of free space $\Phi_j$ in $j$ and a molecule of occupied space $O_{j+1}$ in $j+1$ (cf.\ \Cref{fig:rr}).   
		
		The TRM then falls from writing the kinetics of these reactions using the law of mass action \citep{Feinberg2019} (assuming that the reactions described above happen with some rate $k_j$ that can be time-dependent). As for the coefficients $C_j$ et $q_j$, they are seen as respectively reaction rates between compartments and external intakes or outtakes of reactants (cf.\ \citep{Pereira2021ParameterAD} for details).
		
	\end{remark}
	
	In practice, the modeled road is not infinite, and only a finite number of cells is used to discretize it. Hence, boundary conditions must be supplied to define the evolution of the outer cells of the road. Indeed, assuming that the road is discretized using $N_x$ cells (numbered from $1$ to $N_x$), the recurrence relation~\eqref{eq:trm} takes the form
	\begin{equation}
	\rho_j^{n+1} =\rho_j^n + \left(\tilde{F}_{j-1}^n- \tilde{F}_j^n\right) + q_j^n, \quad j \in \lbrace{1, \dots, N_x}\rbrace,
	\label{eq:recf}
	\end{equation}
	where for the interior cells (cells $2$ to $N_x-1$), the quantities $\tilde{F}_{k}^n$ are defined as the TRM numerical fluxes:
	\begin{equation}
	\tilde{F}_{k}^n=C_{k}^n F(\rho_{k}^n, \rho_{k+1}^n), \quad k \in \lbrace{1, \dots, N_x-1}\rbrace.
	\label{eq:fluxnum}
	\end{equation}
	 For the boundary cells (cells $1$ and $N_x$), the TRM numerical flux would require the quantities $\rho_0^n, \rho_{N_x+1}^n \in [0, \rho_m]$, which are no longer defined. A natural approach would consist in considering these quantities as additional parameters of the model. Then, for instance, the numerical flux $\tilde{F}_{0}^n$ would be given by
	 \begin{equation*}
	 	\tilde{F}_{0}^n=C_{0}^n F(\rho_{0}^n, \rho_{1}^n)= \frac{C_{0}^n\rho_{0}^n}{\rho_m}\bigg(1-\frac{\rho_{1}^n}{\rho_m}\bigg)
	 \end{equation*}
	 and is therefore parameterized only by the product of the independent parameters $C_{0}^n$ and $\rho_{0}^n$ (divided by the maximal density $\rho_m$). Therefore, we propose to directly consider the factor $C_0^n\rho_0^n/\rho_m \in (0,1/2)$ as a parameter of the model, and denote it, with a slight  abuse of notation, by $C_0^n$. Using the same approach for the numerical flux $\tilde{F}_{N_x}^n$, we end up with the following expressions:
	 \begin{equation}
	 \begin{aligned}
	 &\tilde{F}_{0}^n =C_{0}^n\bigg(1-\frac{\rho_{1}^n}{\rho_m}\bigg)=C_{0}^n F(\rho_m, \rho_{1}^n), \\
	 &\tilde{F}_{N_x}^n= C_{N_x}^n \frac{\rho_{N_x}}{\rho_m} =C_{N_x}^n F(\rho_{N_x}, 0),
	 \end{aligned}
	 \label{eq:bc}
	 \end{equation}
	 where $n\in\mathbb{N}_0$. Note that this setting is equivalent to considering that the road is full upstream (i.e. $\rho_0^n=\rho_m$) and empty downstream (i.e. $\rho_{N_x+1}^n=0$). The reaction rates $C_0^n$ and $C_{N_x}^n$ are then used to control the amount of vehicles allowed to enter or exit the road between $t_n$ and $t_{n+1}$.
	
	\begin{remark}
		Note that using \Cref{rem:flux_approx}, the numerical fluxes $F_j^n$ can be assimilated to measurements of the flux of vehicles entering and exiting the road by scaling those by a factor $\rho_m \Delta x/\Delta t$.
	\end{remark}

	\section{Neural networks}
	\label{sec:NN}

	\subsection{Recurrent neural networks}
	\label{sec:num}
	
	Recurrent neural networks (RNN) are a particular type of neural networks designed to take sequential or time-ordered data as inputs, and widely used in applications such as time series prediction, natural language processing, or speech recognition \citep{karpathy2015visualizing}. We now provide a short introduction on these algorithms, but refer the interested reader to \citep{karpathy2015visualizing,sutskever2013training,Goodfellow-et-al-2016,sherstinsky2020fundamentals} for more details on the matter.
	
  Given an input sequence $\bm X =(\bm x^{(1)}, \dots, \bm x^{(\Nseq)})$ (composed of vectors of size $\Nin>0$), a RNN computes a sequence of so-called cell states $\bm S =(\bm s^{(1)}, \dots, \bm s^{(\Nseq)})$ (composed of vectors of size $\Nst>0$) which are turned into a sequence of outputs
	$\bm Y =(\bm y^{(1)}, \dots, \bm y^{(\Nseq)})$ (composed of vectors of size $\Nout>0$). Assuming that an initial state $\bm s^{(0)}$ has been set, the operations performed by the RNN can be summed up by the following recurrence relations:
	\begin{equation}
	\left\lbrace
	\begin{aligned}
	\bm s^{(n)} &=\mathcal{F}_1(\bm s^{(n-1)}, \; \bm x^{(n)}), \\
	\bm y^{(n)} &=\mathcal{F}_2( \bm s^{(n)}, \; \bm x^{(n)}),
	\end{aligned}
	\right.
	\label{eq:rnn}
	\end{equation}
	where $n\in \lbrace 1, \dots, \Nseq$ and  $\mathcal{F}_1, \mathcal{F}_2$ are two nonlinear (vector-valued) functions depending on parameters determined during the training of the RNN. 
	For instance, in the so-called \textit{Simple RNN} (SRNN), the functions $\mathcal{F}_1$ and $\mathcal{F}_2$ are \textit{MultiLayer Perceptrons} (MLPs) (cf.\ \Cref{sec:MLP}). 
	
\begin{figure}
		\centering
		\adjustbox{width=0.65\textwidth}{
			\begin{tikzpicture}

\matrix[column sep=0.65cm, row sep=0.5cm,minimum width=1cm,minimum height=1cm] (m) {
	
	\node[circle,draw,minimum width={width("$\bm h^{(n-1)}$")+5pt},minimum height={width("$\bm h^{(n-1)}$")+5pt}] (ckm1)  {$\bm c^{(n-1)}$};
	
	& 
	
	& \node[rectangle,draw] (times1)  {$\odot$};
	
	& 
	
	& \node[rectangle,draw] (plus)  {$+$};
	
	&
	
	&
	
	&
	
	& \node[circle,draw,minimum width={width("$\bm h^{(n-1)}$")+5pt},minimum height={width("$\bm h^{(n-1)}$")+5pt}] (ck)   {$\bm c^{(n)}$};\\

	& 
	
	& 
	
	& 
	
	& \node[rectangle,draw] (times2)  {$\odot$};
	
	&
	
	&
	
	& \node[rectangle,draw] (sigma)  {$\sigma$};
	
	& \\

	& 
	
	& \node[rectangle,draw] (p1)   {$\mathcal{P}_1$};
	
	& \node[rectangle,draw] (p2)  {$\mathcal{P}_2$};
	
	& 
	
	& \node[rectangle,draw] (p3)  {$\mathcal{P}_3$};
	
	& 
	
	&
	
	& \\

	\node[circle,draw,minimum width={width("$\bm h^{(n-1)}$")+5pt},minimum height={width("$\bm h^{(n-1)}$")+5pt}] (hkm1)  {$\bm h^{(n-1)}$};	
	
	& \node[rectangle,draw] (union)  {$\cup$};
	
	& 
	
	& 
	
	& 
	
	& 
	
	& \node[rectangle,draw] (p4)  {$\mathcal{P}_4$};
	
	& \node[rectangle,draw] (times3)  {$\odot$};
	
	& \node[circle,draw,minimum width={width("$\bm h^{(n-1)}$")+5pt},minimum height={width("$\bm h^{(n-1)}$")+5pt}] (hk)  {$\bm h^{(n)}$};\\

	& \node[circle,draw,minimum width={width("$\bm h^{(n-1)}$")+5pt},minimum height={width("$\bm h^{(n-1)}$")+5pt}] (xk)  {$\bm x^{(n)}$};
	
	& 
	
	& 
	
	& 
	
	& 
	
	& 
	
	& 
	
	& \\		
};

\draw[-latex,very thick] (ckm1) -- (times1);

\draw[-latex,very thick] (times1) -- (plus);

\draw[-latex,very thick] (times2) -- (plus);

\draw[-latex,very thick] (p1) -- (times1);

\draw[-latex,very thick] (plus) -- (ck);

\draw[-latex,very thick] (p2) |- (times2);

\draw[-latex,very thick] (p3) |- (times2);

\draw[-latex,very thick] (p4) -- (times3);

\draw[-latex,very thick] (times3) -- (hk) node (int30) [midway] {};

\draw[-latex,very thick] (sigma) -- (times3);

\draw[-latex,very thick] (ckm1 -| sigma) -- (sigma);

\draw[-latex,very thick]  (union) -- (p4);

\draw[-latex,very thick] (hkm1 -| p1) -- (p1);

\draw[-latex,very thick] (hkm1 -| p2) -- (p2);

\draw[-latex,very thick] (hkm1 -| p3) -- (p3);

\draw[-latex,very thick] (xk) -- (union)  node (int0) [midway] {};
\draw[-latex,very thick] (hkm1) -- (union) node (int0b) [midway] {};

\node (int1) at (int0 -| int0b) {};

\node (int3) at (int1 -| int30) {};

\node (int2) at ([yshift=0.3cm] ckm1.north  -| int1) {};

\node (int4) at (int2 -| int3) {};

\draw[very thick] (int1.center) -- (int3.center) -- (int4.center) -- (int2.center) -- (int1.center);
%\draw (int3.center) -- (int4.center);
%\draw (int3.center) -- (int1.center);
%\draw (int4.center) -- (int2.center);
% \node[circle,draw] (hkm1) [below of= ckm1]  {$\bm h^{(n-1)}$};

\end{tikzpicture}
		}
		\caption{LSTM architecture: $\mathcal{P}_i$, $i\in\lbrace 1, \dots, 4\rbrace$ denote single layer perceptrons,  $\odot$ denotes the component-wise product between two vectors, $+$ denotes the sum of two vectors and $\sigma$ denotes the application of an activation function to the entries of a vector. }	
		\label{fig:lstm}
	\end{figure}
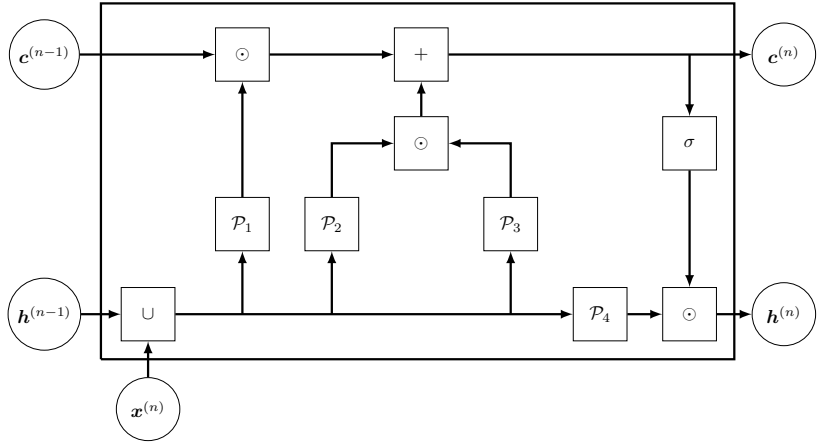

	A shortcoming of RNNs is the problem of vanishing and exploding gradients, which occur when backpropagation is used to compute the gradient of the cost function used when training the algorithm. This can result in a training phase that fails to produce adequate parameters for the model to yield low-error predictions.  The \textit{Long Short-Term Memory} (LSTM) architecture was proposed to avoid this pitfall. In this setting, the state vectors $\bm s^{(n)}$ ($n\in\lbrace 1, \dots, \Nseq\rbrace$) can be seen as composed by two subvectors as $\bm s^{(n)} =(\bm c^{(n)}, \bm h^{(n)})$. The nonlinear functions $\mathcal{F}_1$ and $\mathcal{F}_2$ defining the RNN in~\eqref{eq:rnn} then take the form 
	\begin{equation*}
	\left\lbrace\begin{aligned}
	\bm s^{(n)}
	&=\begin{pmatrix}
	\bm c^{(n)}  \\
	\bm h^{(n)}
	\end{pmatrix}
	=\begin{pmatrix}
	\mathcal{P}_1(\bm h^{(n-1)}, \; \bm x^{(n)}) \odot \bm c^{(n-1)} + \mathcal{P}_2(\bm h^{(n-1)}, \; \bm x^{(n)}) \odot \mathcal{P}_3(\bm h^{(n-1)}, \; \bm x^{(n)}) \\
	\mathcal{P}_4(\bm h^{(n-1)}, \; \bm x^{(n)}) \odot \sigma(\bm c^{(n)}), 
	\end{pmatrix}\\
	\bm y^{(n)} &=\bm h^{(n)},
	\end{aligned}\right. 
	\end{equation*}
	where $n\in \lbrace 1, \dots, \Nseq\rbrace$ and $\mathcal{P}_i$, $i\in\lbrace 1, \dots, 4\rbrace$ denote single layer perceptrons (cf.\ \Cref{sec:MLP}) and $\sigma$ denotes the application of a nonlinear activation function (usually $\tanh$) to all entries of a given vector and $\odot$ denotes the component-wise product between two vectors (cf.\ \Cref{fig:lstm} for a visual representation of an LSTM).
	
	\subsection{The TRM as a RNN}\label{sec:trm_rnn}
	
	Since the TRM can be seen as a forward Euler discretization of a system of ODEs (cf.\ \Cref{rem:ode}), it can also be interpreted as an instance of RNN  for which the successive states of the RNN represent successive states of the discretized system of ODEs \citep{ruthotto2019deep,sherstinsky2020fundamentals} . Hence, taking the notations introduced in the previous subsection, the state vector $\bm s^{(n)}$ of the RNN associated with the TRM is no other than the vector containing the numerical solution of PDE~\eqref{eq:lwr} at the $k$-th time step, i.e.\  $\bm s^{(n)}=(\rho_1^n, \dots, \rho_{N_x}^n)$. As for the input $\bm x^{(n)}$, it is given by the vector containing the reaction rates at the $n$-th time step, i.e.\ $\bm x^{(n)}=(C_0^{n-1}, \dots, C_{N_x}^{n-1})$.
	
	Since our goal is to perform flux predictions, the output of the RNN that we associate to the TRM is taken to be the value of the numerical fluxes $\bm y^{(n)}= (\tilde{F}_0^{n-1}, \dots, \tilde{F}_{N_x}^{n-1})$, as given by the relations \eqref{eq:fluxnum}-\eqref{eq:bc}.  By definition of the TRM recurrence~\eqref{eq:trm}, when omitting the source terms, the corresponding RNN equations~\eqref{eq:rnn} are given, for any $n\in \lbrace 1, \dots, \Nseq\rbrace$, by
	\begin{equation*}
	\left\lbrace
	\begin{aligned}
	\bm s^{(n)} &=\bm s^{(n-1)}+\bm f^{(n-1)}_{0,\dots,N_x-1} - \bm f^{(n-1)}_{1,\dots,N_x}, \\
	\bm y^{(n)} &=\bm f^{(n-1)},
	\end{aligned}
	\right.
	\end{equation*}
	where  $\bm f^{(n-1)}_{0,\dots,N_x-1}$ (resp. $\bm f^{(n-1)}_{1,\dots,N_x}$) denotes the first (resp. last) $N_x$ entries of the vector $\bm f^{(n-1)}$ defined by
	\begin{equation*}
	\renewcommand*{\arraystretch}{1.1}
		\bm f^{(n-1)}=\bm x^{(n-1)} \odot \left(\begin{array}{c}
		1 \\ \hline \\
		\bm s^{(n-1)}/\rho_m \\
		\,
		\end{array}\right)
		\odot
		\left(\begin{array}{c}
		\, \\
		1-\bm s^{(n-1)}/\rho_m \\
		\, \\ \hline 
		1
		\end{array}\right).
	\end{equation*}
	Hence, note that the densities approximated by the TRM  define the internal state of the corresponding RNN, and not its output. We therefore work with an approach similar to state-space models, where the state of the system is the density of vehicles, the dynamics of the system are defined by the TRM (and therefore by extension, the macroscopic traffic flow model~\eqref{eq:lwr}) and the output of the model are the numerical fluxes of the TRM.

	As a RNN, the TRM has a few particularities. First, and as described above, it has no tuning parameters, and its inputs, output and states are interpretable as components of a traffic flow model.  Second, one can note the nonlinearities that appear in the derivation of the TRM are only polynomial, as the states and outputs can be written as polynomial expressions of the inputs and past states.
	
		\subsection{Traffic predictions}\label{sec:pred}
	
	In our setting, we assume that traffic is observed on a stretch of road with no on/off-ramps, for sake of simplicity. The road is discretized into segments of length $\Delta x$ and the flux of vehicles is recorded every $\Delta T >0$ seconds at some locations inside the road. In particular, we assume that the data then consists of measurements of the flux of vehicles at some (or all) of the interfaces between the road segments (or \q{road interfaces} for short). We denote by $N_o$ the number of road interfaces at which the flux is measured and by $N_i \ge N_o$ the total number of road interfaces in the road. In particular, the number $N_s$ of segments used to discretize the road satisfies $N_s=N_i-1$.
	
	We propose to use RNNs to perform short term predictions of the flux of vehicles based on the TRM. More precisely, we start with flux measurements at $N_p$ consecutive time steps $t_{-N_p+1}, \dots, t_{-1},  t_0$ (where $t_{i+1}=t_i+\Delta T$, $i\in\lbrace -N_p+1, \dots, -1\rbrace$) at the $N_o$ observed interfaces of the road and wish to predict the flux of vehicles at time $t_{N_f}=t_0+N_f\Delta T$ at both the observed and unobserved interfaces of the road. Hence we aim at conjointly inpaint and forecast the flux of vehicles along the road.
	
	We propose to use the TRM RNN in \Cref{sec:trm_rnn} to model the evolution of the traffic along the road between $t_{-N_p+1}$ and $t_{N_f}$. We hence consider a TRM defined on the discretized road, and to comply with the CFL condition~\eqref{eq:cfl}, we follow the multilevel approach proposed in \cite{Pereira2021ParameterAD} and consider a time step $\Delta t$ for the TRM given by $\Delta t = \Delta T/P_t$ where $P_t$ is an integer such that
	\begin{equation*}
		P_t > \frac{8\Vert f_m\Vert_{\infty}}{\rho_m} \frac{\Delta T}{\Delta x}.
	\end{equation*}
	Such a choice ensures that $\Delta t$ and $\Delta x$ satisfy the CFL condition~\eqref{eq:cfl}, and therefore the stability of the TRM \cite{Pereira2021ParameterAD,liptak2020traffic}.
	
		\begin{remark}
		Recall then that $\rho_m$ and $\Vert f_m\Vert_{\infty}$ denote respectively the maximal values the density and flux of vehicles can take. The former can be approximated by the ratio between the number of lanes in the road and the average length of the vehicles. As for the latter, note that the speed of vehicles can be identified with the ratio between the flux and the density. Assuming that this speed is bounded by some value $v_{m}$ (for instance $130\,\text{km}/\text{hour}$) and using~\eqref{eq:greenf}, we can upper-bound $\Vert f_m\Vert_{\infty}$ by $v_{m}\rho_m/4$. Hence, we can take $P_t$ to be the smallest integer such that
		\begin{equation*}
		P_t > 2v_{m}\frac{\Delta T}{\Delta x}.
		\end{equation*}	
	\end{remark}

	  Then, the corresponding TRM RNN requires us to specify:
	\begin{itemize}
		\item  an input sequence, taken as the sequence of $(N_f+N_p-1)P_t+1$ vectors, containing the values of the TRM reaction rates (at every road interface) at times $t_{-N_p+1} + k\Delta t$, for $k\in\lbrace 0, \dots, (N_f+N_p-1)P_t\rbrace$;
		\item an initial state, taken as the vector containing the density of vehicles inside each road cell at time~$t_{-N_p+1}$.                             
	\end{itemize}
	The TRM RNN then outputs a sequence of $(N_f+N_p-1)P_t+1$ vectors, containing the values of flux of vehicles at every road interface, between $t_{-N_p+1}$ and $t_{N_f}$: the last element of this sequence gives the desired flux prediction.
	
	In order to specify the input sequence needed by the TRM RNN, we proceed as follows. Starting from the $N_p$ flux observations at times $t_{-N_p+1}, \dots, t_0$ and at the $N_o$ observed road interfaces, we use a LSTM RNN to estimate the reaction rates at the same time stamps and at all of the road interfaces (including the unobserved ones). Namely, we consider a LSTM network whose input sequence is given by the flux observations and whose outputs are used as estimates for the reaction rates at times  $t_{-N_p+1}, \dots, t_0$. We refer to this LSTM network as \q{Extractor} as it is in fact used to extract reaction rates from flux observations. The initial state of this RNN is obtained as the output of a 2-layer MLP taking as input the flux measurement at time $t_{-N_p+1}$.

	Then, the reaction rates at times $t_1, \dots, t_{N_f}$ are predicted using a SRNN whose inputs are vectors of zeros, and with initial state given by the last state of the LSTM RNN (i.e.\ the state of the LSTM RNN at time $t_0$). We refer to this SRNN as \q{Predictor}, as it is in fact used to predict reaction rates  beyond the time scope of the observations (i.e.\ for times greater than $t_0$).
	
	\begin{remark}
		Note that since the LSTM RNN and SRNN are tasked with estimating and predicting reaction rates, their outputs should take values in $(0, 1/2)$, in accordance with the CFL condition (cf.\ \cref{eq:cfl_coef}). This is achieved by choosing the activation functions of these RNNs to sigmoids (hence ensuring that the output is in $(0,1)$) and scaling the output by a factor~$1/2$ before using them.
	\end{remark} 	
	
	Hence, we end up with reaction rates at times $t_{-N_p+1}, \dots, t_0, t_1, \dots,  t_{N_f}$ and at every road interface. We then complete these reaction rates by assuming that the reaction rates stay constant between these time steps, thus allowing us to have reaction rates at any time $t_{-N_p+1} + k\Delta t$, for $k\in\lbrace 0, \dots, (N_f+N_p-1)P_t\rbrace$ (as needed by the TRM RNN). These reaction rates are then used in the TRM RNN described above and yield estimates for the flux of vehicles at the same time steps.
	In particular, since the TRM is nothing but the discretization of the continuous traffic model~\eqref{eq:lwr}, the TRM RNN then recreates, under this traffic model, the traffic dynamics during the time scope of the observations (i.e.\ between $t_{-N_p+1}$ and $t_0$) and extends it to future times (i.e.\ $t_1$ to $t_{N_f}$) to yield predictions.

	In the end, our proposed architecture takes as input a sequence of past flux observations, and outputs a re-estimation of these observations under the assumption that they arise from the traffic flow model~\eqref{eq:lwr}, and then uses this same model to extend the traffic dynamics to future times, hence yielding the flux predictions. 
	In particular, when the observations consist in noisy (raw) flux measurements, 
	their re-estimation by the TRM RNN can be seen as a smoothing of the measurements. This smoothing is a consequence of the smooth nature of the TRM as a finite volume discretization of a PDE, and was already  noted in \citep{Pereira2021ParameterAD}. Hence, it has a physical interpretation as resulting from the traffic flow model~\eqref{eq:lwr}, and is integrated to the learning process.
	We sum up in \Cref{fig:archi_det} the architecture used for flux predictions.

	\begin{figure}
	\centering
	\adjustbox{width=\textwidth}{
		\begin{tikzpicture}

\matrix[column sep=5ex, row sep=5ex, minimum height=10ex, align=center] (m) {
	
	&
	
	&
	
	\node[ellipse] (d1) {\adjustbox{scale=1.5}{\usebox{\obs}}};
	
	& 	\node[ellipse] (d2) {\adjustbox{scale=1.5}{\usebox{\obs}}};

	& 	\node[ellipse] (d3) {\adjustbox{scale=1.5}{\usebox{\obs}}};
		
	& 	\node (dd) {$\cdots$};	

	& 	\node[ellipse] (df) {\adjustbox{scale=1.5}{\usebox{\obs}}};	
	
	& 
	
	& 
	
	& 
	
	& 
	
	\\
	
	& \node[rectangle,draw] (mlp1) {MLP 1};
	
	& \node[rectangle,draw,lstm] (l1) {LSTM};

	& 	\node[rectangle,draw,lstm] (l2) {LSTM};
	
	& 	\node[rectangle,draw,lstm] (l3) {LSTM};
	
	& 	\node (ld) {$\cdots$};	
	
	& 	\node[rectangle,draw,lstm] (lf) {LSTM};	
	
	& 
	
	& 
	
	& 
	
	& 
	
	\\	
	
	&
	
	& \node[ellipse] (re1) {\adjustbox{scale=1.5}{\usebox{\all}}};
	
	& 	\node[ellipse] (re2) {\adjustbox{scale=1.5}{\usebox{\all}}};
	
	& 	\node[ellipse] (re3) {\adjustbox{scale=1.5}{\usebox{\all}}};
	
	& 	\node (red) {$\cdots$};	
	
	& 	\node[ellipse] (ref) {\adjustbox{scale=1.5}{\usebox{\all}}};

	& 	\node[rectangle,draw,srnn] (s1) {SRNN};
	
	& 	\node[rectangle,draw,srnn] (s2) {SRNN};
	
	& 	\node (sd) {$\cdots$};	
	
	& 	\node[rectangle,draw,srnn] (sf) {SRNN};		
	
	\\
	
	&
	
	&
	
	& 
	
	& 
	
	& 
	
	&

	& 	\node[ellipse] (rp1) {\adjustbox{scale=1.5}{\usebox{\all}}};
	
	& 	\node[ellipse] (rp2) {\adjustbox{scale=1.5}{\usebox{\all}}};
	
	& 	\node (rpd) {$\cdots$};	
	
	& 	\node[ellipse] (rpf) {\adjustbox{scale=1.5}{\usebox{\all}}};	
	
	\\
	
	\node[rectangle,draw] (mlp2) {MLP 2};
	
	&
	
	&	\node[rectangle,draw,trm] (te1) {TRM};
	
	& 	\node[rectangle,draw,trm] (te2) {TRM};
	
	& 	\node[rectangle,draw,trm] (te3) {TRM};
	
	& 	\node (ted) {$\cdots$};	
	
	& 	\node[rectangle,draw,trm] (tef) {TRM};

	& 	\node[rectangle,draw,trm] (tp1) {TRM};
	
	& 	\node[rectangle,draw,trm] (tp2) {TRM};
	
	& 	\node (tpd) {$\cdots$};	
	
	& 	\node[rectangle,draw,trm] (tpf) {TRM};		
	
	\\
	
	&
	
	& \node[rectangle] (e1) {\adjustbox{scale=1.5}{\usebox{\all}}};
	
	& 	\node[rectangle] (e2) {\adjustbox{scale=1.5}{\usebox{\all}}};
	
	& 	\node[rectangle] (e3) {\adjustbox{scale=1.5}{\usebox{\all}}};
	
	& 	\node (ed) {$\cdots$};	
	
	& 	\node[rectangle] (ef) {\adjustbox{scale=1.5}{\usebox{\all}}};

	& 	\node[rectangle] (p1) {\adjustbox{scale=1.5}{\usebox{\all}}};
	
	& 	\node[rectangle] (p2) {\adjustbox{scale=1.5}{\usebox{\all}}};
	
	& 	\node (pd) {$\cdots$};	
	
	& 	\node[rectangle] (pf) {\adjustbox{scale=1.5}{\usebox{\all}}};		
	
	\\	
	
	&

	& \node[rectangle] (ne1) {$t_{-N_p+1}$};

	& 	\node[rectangle] (ne2) {$t_{-N_p+2}$};
	
	& 	\node[rectangle] (ne3) {$t_{-N_p+3}$};
	
	& 	\node (ned) {$\cdots$};	
	
	& 	\node[rectangle] (nef) {$t_{0}$};

	& 	\node[rectangle] (np1) {$t_{1}$};
	
	& 	\node[rectangle] (np2) {$t_{2}$};
	
	& 	\node (npd) {$\cdots$};	
	
	& 	\node[rectangle] (npf) {$t_{N_f}$};		
	
	\\	
};

\draw[-latex,thick] ([yshift=3ex,xshift=-5ex] ne1.west) -- ([yshift=3ex,xshift=5ex] npf.east);
\node at ([yshift=3ex,xshift=6ex] npf.east) {$t$};
\draw[thick] ([yshift=4ex] ne1.center) -- ([yshift=2ex] ne1.center);
\draw[thick] ([yshift=4ex] ne2.center) -- ([yshift=2ex] ne2.center);
\draw[thick] ([yshift=4ex] ne3.center) -- ([yshift=2ex] ne3.center);
\draw[thick] ([yshift=4ex] nef.center) -- ([yshift=2ex] nef.center);
\draw[thick] ([yshift=4ex] np1.center) -- ([yshift=2ex] np1.center);
\draw[thick] ([yshift=4ex] np2.center) -- ([yshift=2ex] np2.center);
\draw[thick] ([yshift=4ex] npf.center) -- ([yshift=2ex] npf.center);

\draw[-latex,thick] (d1) -- (l1);
\draw[-latex,thick] (d2) -- (l2);
\draw[-latex,thick] (d3) -- (l3);
\draw[-latex,thick] (df) -- (lf);

\draw[-latex,thick] (l1) -- (re1);
\draw[-latex,thick] (l2) -- (re2);
\draw[-latex,thick] (l3) -- (re3);
\draw[-latex,thick] (lf) -- (ref);

\draw[-latex,thick,lstm] (l1) -- (l2);
\draw[-latex,thick,lstm] (l2) -- (l3);
\draw[-latex,thick,lstm] (l3) -- (ld);
\draw[-latex,thick,lstm] (ld) -- (lf);

\draw[-latex,thick] (ref) -- (s1);

\draw[-latex,thick,srnn] (s1) -- (s2);
\draw[-latex,thick,srnn] (s2) -- (sd);
\draw[-latex,thick,srnn] (sd) -- (sf);

\draw[-latex,thick] (s1) -- (rp1);
\draw[-latex,thick] (s2) -- (rp2);
\draw[-latex,thick] (sf) -- (rpf);

\draw[-latex,thick] (re1) -- (te1);
\draw[-latex,thick] (re2) -- (te2);
\draw[-latex,thick] (re3) -- (te3);
\draw[-latex,thick] (ref) -- (tef);

\draw[-latex,thick] (rp1) -- (tp1);
\draw[-latex,thick] (rp2) -- (tp2);
\draw[-latex,thick] (rpf) -- (tpf);

\draw[-latex,thick,trm] (te1) -- (te2);
\draw[-latex,thick,trm] (te2) -- (te3);
\draw[-latex,thick,trm] (te3) -- (ted);
\draw[-latex,thick,trm] (ted) -- (tef);
\draw[-latex,thick,trm] (tef) -- (tp1);
\draw[-latex,thick,trm] (tp1) -- (tp2);
\draw[-latex,thick,trm] (tp2) -- (tpd);
\draw[-latex,thick,trm] (tpd) -- (tpf);

\draw[-latex,thick] (te1) -- (e1);
\draw[-latex,thick] (te2) -- (e2);
\draw[-latex,thick] (te3) -- (e3);
\draw[-latex,thick] (tef) -- (ef);
\draw[-latex,thick] (tp1) -- (p1);
\draw[-latex,thick] (tp2) -- (p2);
\draw[-latex,thick] (tpf) -- (pf);

\draw[-latex,thick] (d1) -| (mlp1);
\draw[-latex,thick] (mlp1) -- (l1);

\draw[-latex,thick] (d1) -| (mlp2);
\draw[-latex,thick] (mlp2) -- (te1);

\node[inner sep=4pt,rounded corners=2ex,draw,align=center,dashed,data] (data) [fit=(d1) (df)] {};
\node[data] (data_txt) at ([yshift=2ex] data.north){(Incomplete) Flux measurements};

\node[inner sep=4pt,rounded corners=2ex,draw,align=center,dashed,pcol] (re) [fit=(re1) (ref)] {};
\node[black] (re_txt) at ([xshift=-3.5em] re.west){\begin{tabular}{c}
	Estimated \\[-2ex]  reaction rates
	\end{tabular}};

\node[inner sep=4pt,rounded corners=2ex,draw,align=center,dashed,pcol] (rp) [fit=(rp1) (rpf)] {};
\node[black] (rp_txt) at ([xshift=3.5em] rp.east){\begin{tabular}{c}
	Predicted \\[-2ex]  reaction rates
	\end{tabular}};

\node[inner sep=4pt,rounded corners=2ex,draw,align=center,dashed,ecol] (smooth) [fit=(e1) (ef)] {};
\node[ecol] (smooth_txt) at ([yshift=-2ex] smooth.south){Flux reestimation};

\node[inner sep=4pt,rounded corners=2ex,draw,align=center,dashed,pcol] (pred) [fit=(p1) (pf)] {};
\node[pcol] (pred_txt) at ([yshift=-2ex] pred.south){Flux predictions};

\node[inner sep=9pt,rectangle,draw,align=center,dashed,lstm] (lstm) [fit=(l1) (lf)] {};
\node[lstm] (lstm_txt) at ([xshift=7ex] lstm.east){Extractor};

\node[inner sep=9pt,rectangle,draw,align=center,dashed,srnn] (srnn) [fit=(s1) (sf)] {};
\node[srnn] (srnn_txt) at ([xshift=7ex] srnn.east){Predictor};

\node at ($(e1)!0.5!(e2)$) {$\dots$};
\node at ($(e2)!0.5!(e3)$) {$\dots$};
\node at ($(p1)!0.5!(p2)$) {$\dots$};

\node[inner sep=9pt,rectangle,draw,align=center,dashed,trm] [fit=(te1) (tpf)] {};

\end{tikzpicture}
	}
	\caption{Detail of the architecture used for flux predictions. The circles stand for road interfaces and green if an observation or estimate is available, and  black otherwise. \\ The data, which consists a sequence of flux observations, is fed to a LSTM RNN (the Extractor) tasked with estimating the reaction rates that gives rise to the flux observations. These reaction rates are then extended to future time steps using a second RNN, the Predictor. All these reaction rates then serve as input of a TRM RNN which produces a smoothed version of the observed traffic states, and extends them to future time steps. The blocks MLP 1 and MLP 2 correspond to the multilayer perceptrons used to initialize respectively the Extractor and the TRM.}
	\label{fig:archi_det}
\end{figure}

	\begin{table}
		\centering
		\adjustbox{width=\textwidth}{
		\begin{tabular}{|c||c|c|c||c|}
			\hline
			   & Input size &  State size & Output size & Number of parameters \\
			\hline
			MLP 1 & $N_o$ & $-$ & $2 N_i$ & $2N_i(2N_i+N_o+2)$ \\
			\hline
			Extractor & $N_p \times N_o$ & $2 N_i$ & $ N_p \times N_i$ & $4N_i(N_i+N_o+1)$ \\
			\hline
			MLP 2 & $N_o$ & $-$ & $N_i-1$ & $(N_i-1)(N_i+N_o+1)$ \\
			\hline
			Predictor & $N_p \times N_i$ & $N_i$ & $N_f \times N_i$ & $3N_i(N_i+1)$ \\
			\hline
			TRM & $\big((N_f + N_p-1)P_t+1\big)\times N_i$ & $N_i-1$ & $ \big((N_f + N_p-1)P_t+1\big)\times N_i$ & $0$ \\
			\hline
			Total & $-$ & $-$ & $-$ & $\begin{aligned}
			12 &N_i^2 +7N_iN_o \\&+11N_i-N_o-1
			\end{aligned}$ \\
			\hline
		\end{tabular}
	}
	\vspace{1ex}
		\caption{Sizes and number of parameters.} \label{tab:param}
	\end{table}
	
	The number of free parameters to learn, as well as the size of the input, output and (when applicable) state vectors or sequences of the different networks involved in our prediction approach are presented in \Cref{tab:param}. These parameters are determined through training, in a supervised learning setting: hence, a cost function evaluating the discrepancy between \q{ideal} outputs of the network and the actual outputs obtained is minimized to determine a set of optimal parameters.
	
	In practice, we perform training as follows. We assume that we have at our disposal a \q{long}\footnote{\q{Long} here can be understood as having a total time scope for the observation much larger than $(N_p+N_f)\Delta t$.} history of flux observations. From these observations, we can form a large set of so-called training examples, which consist of pairs $(\bm F, \tilde{\bm F})$ where 
	\begin{itemize}
		\item $\bm F$ is a sequence of $N_p$ consecutive flux observations denoted by $\bm F=(\bm f_{-N_p+1}, \dots, \bm f_{0})$ and corresponding to some consecutive observation times $t_{-N_p+1}, \dots, t_0$;
		\item ${\bm \Phi}$ is a sequence of $N_p+N_f$ consecutive flux observations denoted by $\bm \Phi=(\bm f_{-N_p+1}, \dots, \bm f_{N_f})$ and corresponding to some consecutive observation times $t_{-N_p+1}, \dots, t_0, \dots, t_{N_f}$; note in particular that the first $N_p$ elements of $\bm \Phi$ are the same as the elements of $\bm F$.
	\end{itemize}
	In an example $(\bm F, \bm \Phi)$, $\bm F$ is used as an input of the flux prediction network $\mathcal{N}$ that we are training, and $\bm \Phi$ is used as the ideal output we would like to get from $\mathcal{N}$. 
	
	Let us denote by $\widehat{\bm \Phi}= \mathcal{N}(\bm F, \bm \theta)=(\widehat{\bm f}_{-N_p+1}, \dots, \widehat{\bm f}_{N_f})$ the output of $\mathcal{N}$ with a set of parameters $\bm \theta$ and when inputting $\bm F$, and let us consider a set of $\Nex$ examples $\lbrace (\bm F^{(1)}, \bm \Phi^{(1)}), \dots, (\bm F^{(\Nex)}, \bm \Phi^{(\Nex)})\rbrace$. We determine the optimal set of parameters $\bm \theta$ associated with these examples by minimizing the cost function $L$ given by
	\begin{equation*}
		L(\bm\theta)= 
		\frac{1}{\Nex}\sum_{j=1}^{\Nex} \frac{1}{N_p}\sum_{k=-N_p+1}^0 \big\Vert \widehat{\bm f}^{(j)}_k-\bm f^{(j)}_k\big\Vert^2 
		+ \frac{1}{\Nex}\sum_{j=1}^{\Nex}\frac{1}{N_f}\sum_{l=1}^{N_f} \big\Vert \widehat{\bm f}^{(j)}_l-\bm f^{(j)}_l\big\Vert^2  
		+ R(\bm \theta),
	\end{equation*}
	where $\Vert\cdot\Vert$ stands for the Euclidean norm of vectors, and $R(\bm\theta)$ is a regularization term. Note that the first two terms defining this cost functions have very particular roles: the first one assesses the capacity of the TRM to recreate the dynamics of the observations, and the second one assesses its capacity to extend it to make predictions. As for the regularization term, its role is to counteract the tendency the TRM might have to overfit the data through physically unsound reaction rates, as investigated in~\cite{Pereira2021ParameterAD}. As proposed in~\cite{Pereira2021ParameterAD}, a regularization term limiting the spatial variations of these reaction rates can be used for this purpose: we hence choose $R(\bm\theta)$ to be defined as
	\begin{equation*}
		R(\bm\theta)=\frac{1}{N_p+N_f}\sum_{n=-N_p+1}^{N_f}\frac{1}{N_x}\sum_{i=0}^{N_x-1} \big( C_{i+1}^{n} - C_{i}^n\big)^2.
	\end{equation*}

	\section{Numerical experiments}
	\label{sec:num}
	
	\subsection{Setup}
	The code used in this section was implemented in Python, using Tensorflow and Keras (2.4.0). We use a dataset which consists of measurements of flux made by detectors placed along a stretch of freeway A12 in the Netherlands (courtesy of Rijkswaterstaat -- Centre for Transport and Navigation). This data was collected (almost) every day from January 5, 2006 to May 19, 2006. Each day, the flux and speed of vehicles at each station were recorded every minute, for a total duration of 5 hours.

			\begin{figure}[ht]
		\centering
		\adjustbox{width=0.7\textwidth}{
			\begin{tikzpicture}[x=10cm]

\foreach \x in {0, 0.04, 0.08, 0.12, 0.17, 0.21, 0.29, 0.37, 0.44, 0.49, 0.57,
	0.63, 0.69, 0.77, 0.85, 0.93, 1}{
	\draw[line width=1.5,myred] (\x,4.85) -- (\x,5.15);
}

\draw[very thick,-latex] (-0.1,5) -- (1.1,5);
%\node at ([xshift=27] n9.center) {m};
%
%\foreach \x in {1,2,...,9}{
%	\draw[line width=2,myred] (n\x.north) -- (n\x.south);
%}

\foreach \x in {0 ,  0.0183, 0.0365, 0.0548, 0.0731, 0.0914, 0.1096, 0.1279,
	0.1462, 0.1644, 0.1827, 0.201 , 0.2192, 0.2375, 0.2558, 0.2741,
	0.2923, 0.3106, 0.3289, 0.3471, 0.3654, 0.3837, 0.4019, 0.4202,
	0.4385, 0.4568, 0.475 , 0.4933, 0.5116, 0.5298, 0.5481, 0.5664,
	0.5847, 0.6029, 0.6212, 0.6395, 0.6577, 0.676 , 0.6943, 0.7125,
	0.7308, 0.7491, 0.7674, 0.7856, 0.8039, 0.8222, 0.8404, 0.8587,
	0.877 , 0.8952, 0.9135, 0.9318, 0.9501, 0.9683, 0.9866}{
	\draw[line width=0.75,dashed] (\x,4.35) -- (\x,3.85);
}

\foreach \x/\i in {0/0 ,  0.0365/2, 0.0731/4, 0.1279/7, 0.1644/9, 0.2192/12, 0.2923/16, 0.3654/20,
	0.4385/24, 0.4933/27, 0.5664/31, 0.6395/35, 0.6943/38, 0.7674/42, 0.8587/47, 0.9318/51,
	0.9866/54}{
	\node at (\x, 3.55) {\i};
	\draw[line width=0.75,myred] (\x,4.35) -- (\x,3.85);
}

\draw[very thick] (-0.1,4.35) -- (1.1,4.35);
\draw[very thick] (-0.1,3.85) -- (1.1,3.85);

%
%\foreach \x in {3,4}{
%	\foreach \y in {0,1,...,11}{
%	\node (nb\x\y) at (\y/11,\x)  {};
%	}
%}
%\foreach \x in {0,1,...,11}{
%	\draw[line width=1,dashed] (nb3\x.center) -- (nb4\x.center);
%}
%\foreach \x in {0,1,2,3,4,5,7,9,11}{
%	\draw[line width=1.5,myred] (nb3\x.center) -- (nb4\x.center);
%}
%\foreach \x in {3,4}{
%	\draw[line width=1.5] ([xshift=-10] nb\x0.center) -- ([xshift=20] nb\x11.center);
%}
%
%
%\foreach \x in {0,1,...,11}{
%	\node at (\x/11,2.5) {\x};
%}
%

%\foreach \x in {1,2}{
%	\foreach \y in {0,1,...,11}{
%		\node (nb\x\y) at (\y/11,\x)  {};
%	}
%}
%\foreach \x in {0,1,...,11}{
%	\draw[line width=1,dashed] (nb1\x.center) -- (nb2\x.center);
%}
%\foreach \x in {0,1,4,5,7,9,11}{
%	\draw[line width=1.5,myred] (nb1\x.center) -- (nb2\x.center);
%}
%\foreach \x in {2,3}{
%	\draw[line width=1.5,myred,dashed] (nb1\x.center) -- (nb2\x.center);
%}
%\foreach \x in {1,2}{
%	\draw[line width=1.75] ([xshift=-10] nb\x0.center) -- ([xshift=20] nb\x11.center);
%}

\end{tikzpicture}
		}
		\caption{Road and detectors. The red ticks on the horizontal axis mark the actual position of the detectors on the road. Below this axis, we represent the road discretization we consider, obtained by dividing the road into segments of $150\,\text{m}$: we use continuous red lines for the road interfaces for which data are available, and black dashed lines for the others.}	
		\label{fig:road}
	\end{figure}
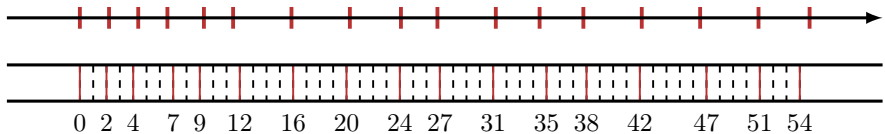
	
	 In the numerical experiment, we consider only a subsection of the road of length $\approx 8200\, \text{m}$. This stretch of road  and the position of the detectors is represented in \Cref{fig:road}, where we can notice that the detectors are approximately separated by (multiples of) $150\,\text{m}$. This allows us to discretize the road into segments of length $\Delta x=150\,\text{m}$ and have the detectors located at the interfaces of these segments (cf.\ \Cref{fig:road}). Note in particular that the considered stretch of road contains an off-ramp (which starts at the location of the interface~$24$) and an on-ramp (which ends at the location of the interface~$31$). Nevertheless, we do not explicitly include them in the TRM as we have no information on the in- and out-fluxes of vehicles they generate.

		Starting from the Dutch motorway dataset, we create a training dataset from the measurements made in February, March and April (totaling 84 days) and then use the measurements made in the first 5 days of May to create a validation dataset (used to choose the hyperparameters of the algorithm) and finally the measurements made on the 5 days afer as a test set (used to check the performances of the algorithm). These measurements are used as is, without any smoothing or preprocessing (besides a spatial linear interpolation when a sensor stopped working). Besides, while training the algorithm, we hide the data from the detectors marked as $4$ and $7$ in \Cref{fig:road}, in order to assess the capacity of the algorithm to complete and predict the flux at unobserved interfaces. In the end, we have $N_o=15$ observed interfaces (those marked $0, 2, 9, 12, 16, 20, 24, 27, 31, 35, 38, 42, 47, 51$ and $54$ in \Cref{fig:road}), for a total of $N_i=55$ road interfaces. 
		 
		 We consider the problem of predicting the flux at a road interface up to $10$~minutes in the future. Taking the notations introduced in \Cref{sec:pred}, we therefore consider the architecture yielding a prediction $N_f=10$ time steps in the future. The algorithm will then return both a smooth estimate of the most recent measurement it received as an input, as well as predictions of what the flux might be $1,2, \dots, 10$ time steps in the future (one time step being equal to $1$ minute). 
		 
		 \subsection{Analysis of the results}
		 
		  In order to determine the number of past observations $N_p$, we perform a quick grid search: we train the algorithm with $N_p \in \lbrace 5,7,10,12,15,18,20\rbrace$ and choose the value of $N_p$ yielding the smallest root-mean-square error (RMSE) and mean absolute percentage error\footnote{The mean absolute percentage error between a set of $n$ measurements $y^{(1)}, \dots,y^{(n)}$  and some associated predictions $\hat y^{(1)}, \dots, \hat y^{(n)}$ is defined as: $\text{MAPE} = (1/n)\sum_{k=1}^n \vert \hat y^{(k)} - y^{(k)}\vert/\vert y^{(k)}\vert $.} (MAPE) over the validation dataset. These errors are reported in \Cref{tab:grid_search}. Based on these results, we take  $N_p=15$ as this values yields small values of both RMSE and MAPE. Once that the number of past observations is chosen, we apply the associated training algorithm to the test dataset to assess its performance. In particular, all subsequent errors will be given in terms of MAPE in order to ease the comparison with existing methods carried out in \Cref{sec:compar_ex}.

	\begin{table}
		 	\centering
		 	\adjustbox{width=0.9\textwidth}{
		 	\begin{tabular}{*{8}{C{2cm}}}
\toprule
{} &   $N_p=5$ &   $N_p=7$ &  $N_p=10$ &  $N_p=12$ &  $N_p=15$ &  $N_p=18$ &  $N_p=20$ \\ \midrule
RMSE &  7.14e-03 &  6.94e-03 &  6.81e-03 &  7.00e-03 &  6.83e-03 &  7.09e-03 &  6.92e-03 \\ \hline
MAPE &   13.94\% &    13.5\% &   13.68\% &   13.65\% &   13.34\% &   15.33\% &   14.86\% \\ \bottomrule
\end{tabular}

		 	}
		 	\vspace{1ex}
		 	\caption{Grid search results: prediction RMSE and MAPE on the validation dataset when predicting up to $N_f=10$ time steps ahead using $N_p$ past observations.}
		 	\label{tab:grid_search}
	\end{table}

\begin{figure}
		\begin{subfigure}{0.48\textwidth}
			\centering
			\includegraphics[width=\textwidth]{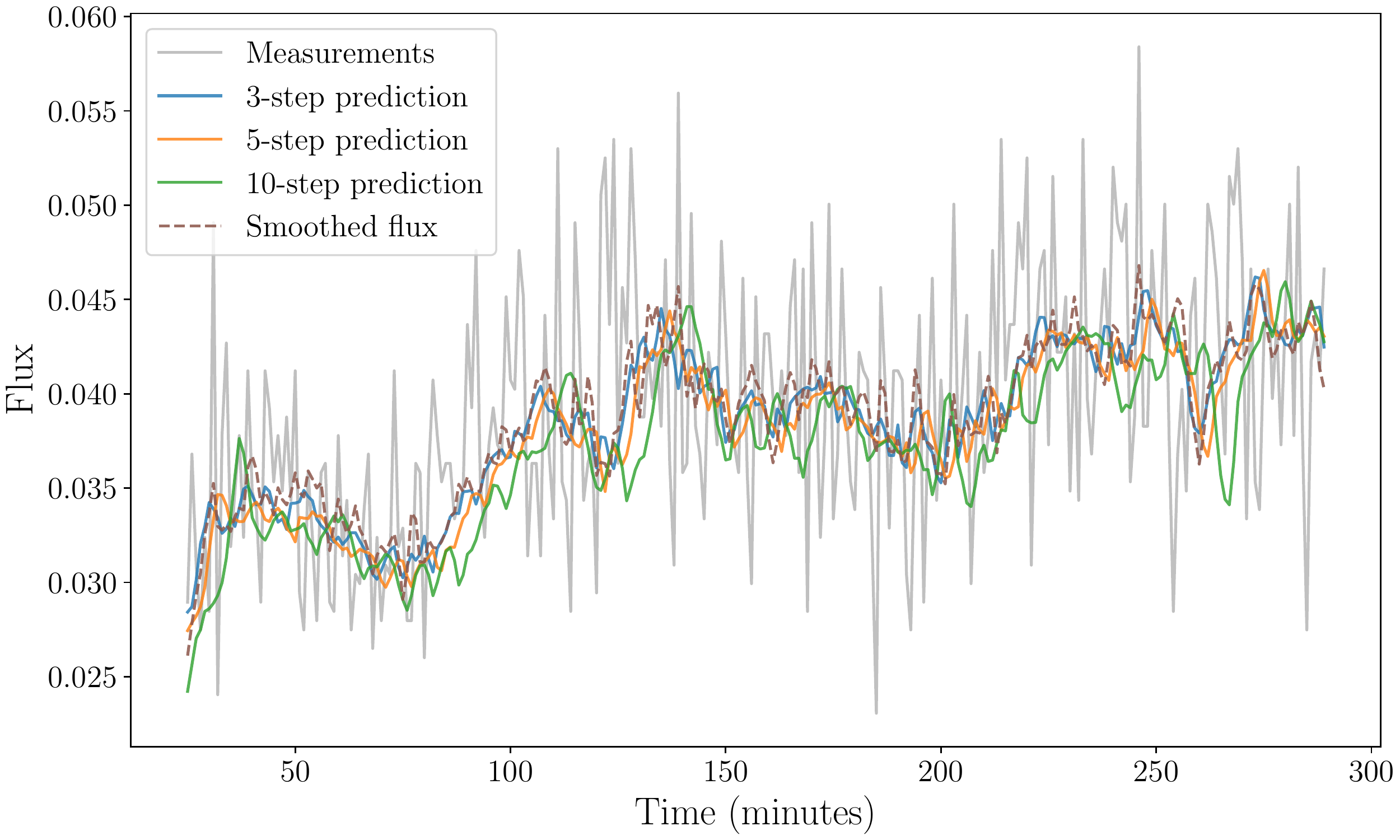}
			\vspace{-1.5em}
			\caption{Road interface 0.}\label{fig:ex_pred_1}
		\end{subfigure}%
		\hfill
		\begin{subfigure}{0.48\textwidth}
			\centering
			\includegraphics[width=\textwidth]{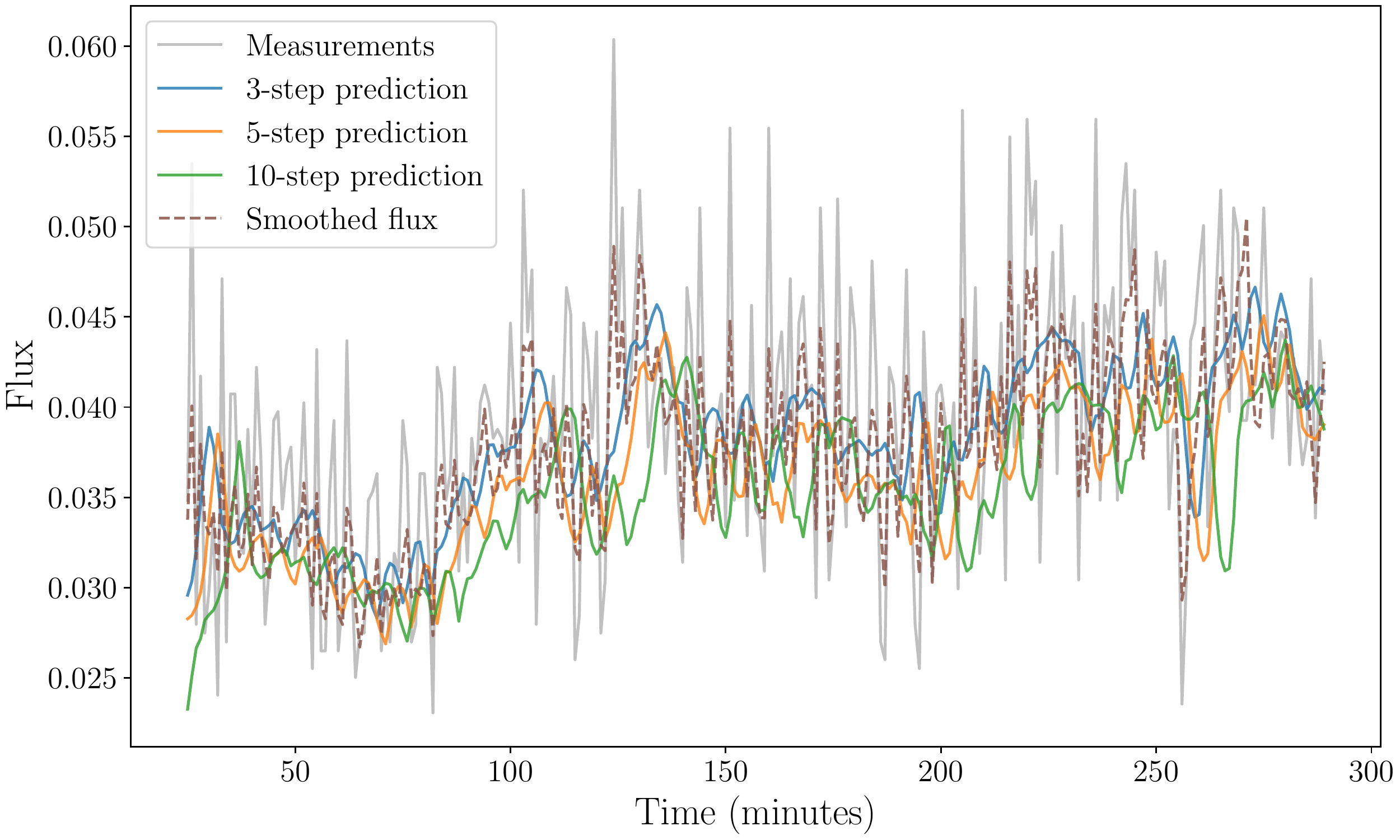}
			\vspace{-1.5em}
			\caption{Road interface 20.}\label{fig:ex_pred_2}
		\end{subfigure}\\
	\vspace{1em}
		\begin{subfigure}{0.48\textwidth}
			\centering
			\includegraphics[width=\textwidth]{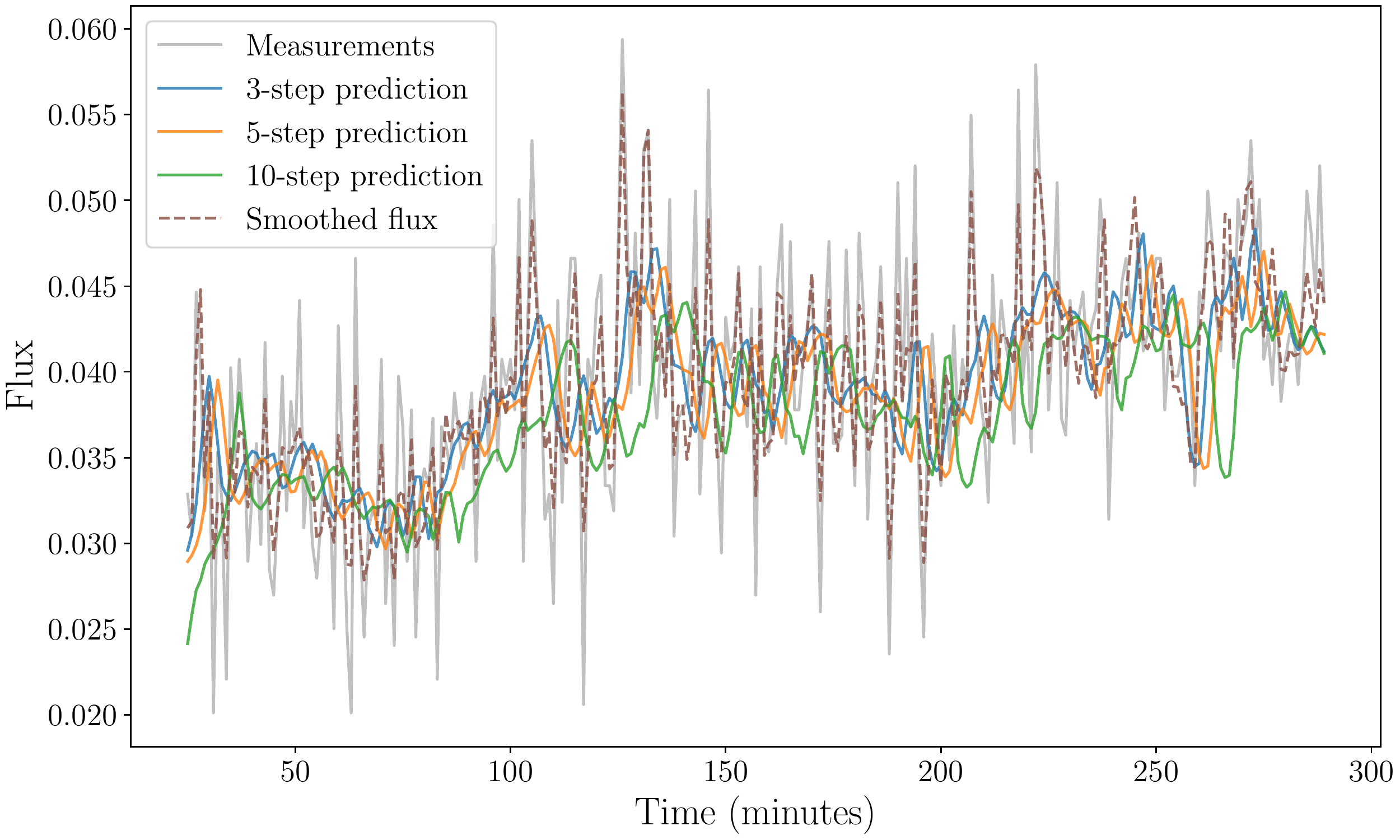}
			\vspace{-1.5em}
			\caption{Road interface 38.}\label{fig:ex_pred_3}
		\end{subfigure}%
	\hfill
	\begin{subfigure}{0.48\textwidth}
		\centering
		\includegraphics[width=\textwidth]{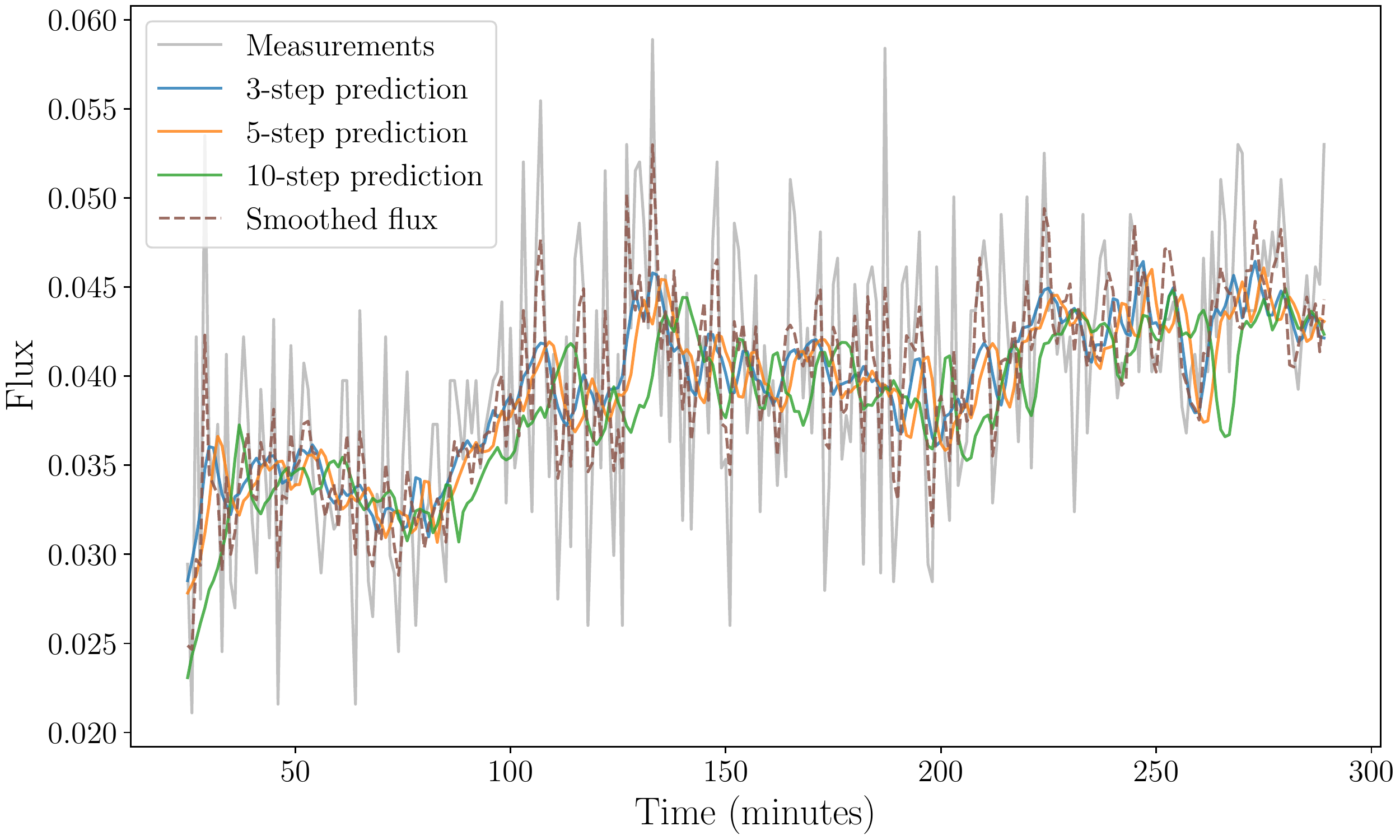}
		\vspace{-1.5em}
		\caption{Road interface 54.}\label{fig:ex_pred_4}
	\end{subfigure}\\
	\vspace{1em}
		\begin{subfigure}{0.48\textwidth}
			\centering
			\adjustbox{width=\textwidth}{
				\begin{tabular}{*{3}{C{3cm}}}
\toprule
{} &  TRM Prediction  &  Last known measurement \\ \midrule
3-step prediction  &          12.98\% &                 17.69\% \\ \hline
5-step prediction  &          13.45\% &                 17.65\% \\ \hline
10-step prediction &          14.16\% &                 18.04\% \\ \bottomrule
\end{tabular}

			}
			\vspace{1ex}
			\caption{Prediction MAPE obtained when using our approach (TRM) and when repeating the last known measurement.}\label{tab:err_pred_obs}
		\end{subfigure}%
		\caption{Measured, smoothed and predicted fluxes on various road interfaces on May 10th, and overall MAPE between predictions and measurements on the test dataset (using only observed road interfaces).}
	\end{figure}

	First, we look at how good the predictions $\psa$, $\psb$ and $\psc$ minutes ahead are. Examples of such predictions are presented in \Cref{fig:ex_pred_1,fig:ex_pred_3,fig:ex_pred_2,fig:ex_pred_4}, where we show, for various road interfaces along the road, a comparison between the $\psa$-, $\psb$- and $\psc$-step predictions and the corresponding measurements obtained on May 10th (which is one of the days used in the test dataset). We also compute, over all the measurements in the test dataset and considering only the road interfaces used in the training phase, the MAPE between predictions and measurements (also called prediction MAPE). We compare them to the MAPE obtained when we use the last known measurement (i.e.\ the measurement at time $t_0$) as a predictor for the flux in the future (i.e.\ at time $t_{\psa}, t_{\psb}$ or $t_{\psc}$). These errors are reported in \Cref{tab:err_pred_obs}. We note that the TRM predictions show an improvement of $22\%$ to $27\%$ compared to the case where the last known measurement is used as predictor.
	
\begin{figure}
		\centering
		\begin{subfigure}{\textwidth}
		\includegraphics[width=\textwidth]{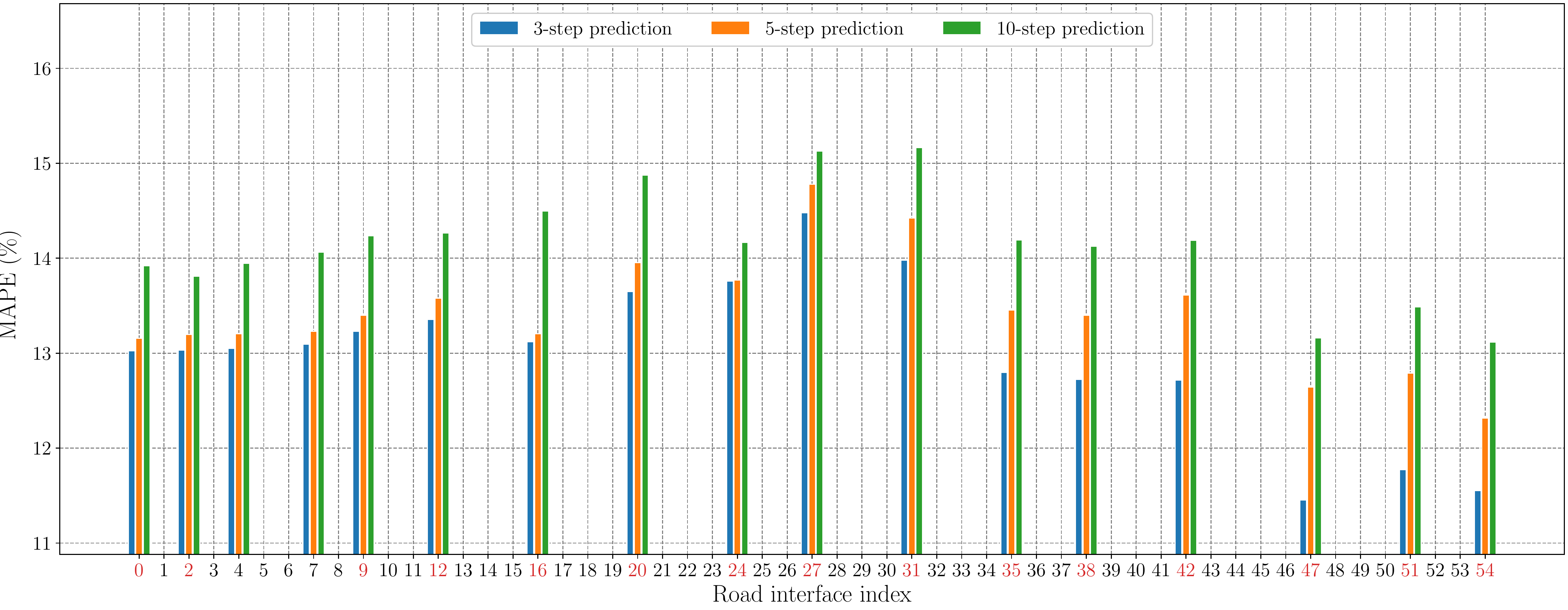}
	\vspace{-1.5em}
	\caption{Prediction MAPE on each road interface when predicting $N_f$ time steps ahead. The indices marked in red correspond to the observed interfaces, and those in black to the hidden interfaces.}\label{fig:err_pred_space}
		\end{subfigure}\\
	\vspace{1em}
		\begin{subfigure}{0.48\textwidth}
			\centering
			\adjustbox{width=\textwidth}{
				\begin{tabular}{*{3}{C{3cm}}}
\toprule
{} &  On observed interfaces  &  On hidden interfaces \\ \midrule
3-step prediction  &                  12.98\% &               13.07\% \\ \hline
5-step prediction  &                  13.45\% &               13.22\% \\ \hline
10-step prediction &                  14.16\% &               14.01\% \\ \bottomrule
\end{tabular}

			}
			\vspace{1ex}
			\caption{Prediction MAPE on the observed and hidden interfaces.}\label{tab:err_pred_compar_obs_unobs}
		\end{subfigure}	 \\
	\vspace{1em}
	\begin{subfigure}{0.48\textwidth}
		\includegraphics[width=\textwidth]{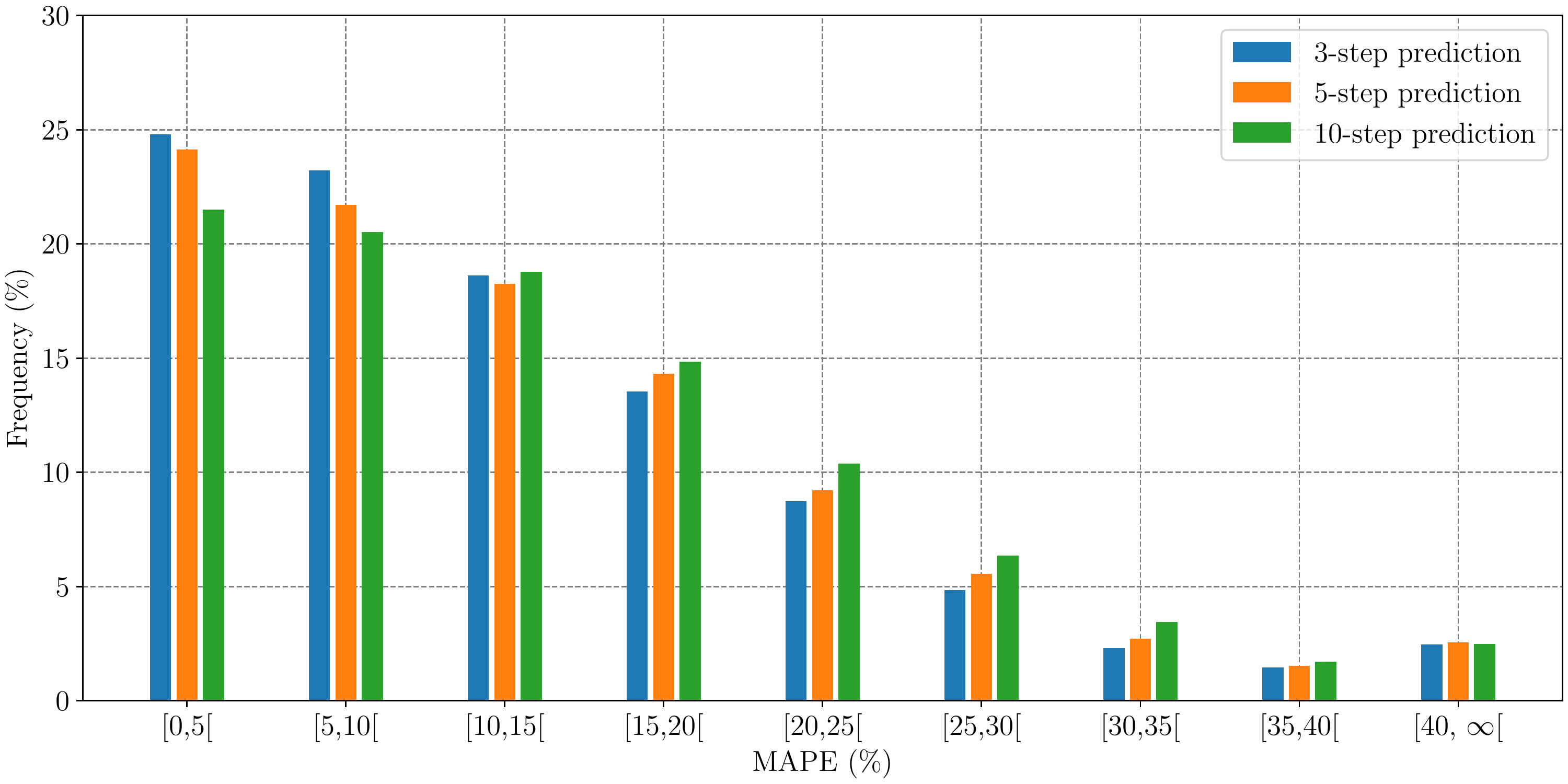}
		\vspace{-1.5em}
		\caption{Distribution of MAPE values on observed interfaces.}\label{fig:dist_err_pred_compar_obs}
	\end{subfigure}%
	\hfill
	\begin{subfigure}{0.48\textwidth}
		\includegraphics[width=\textwidth]{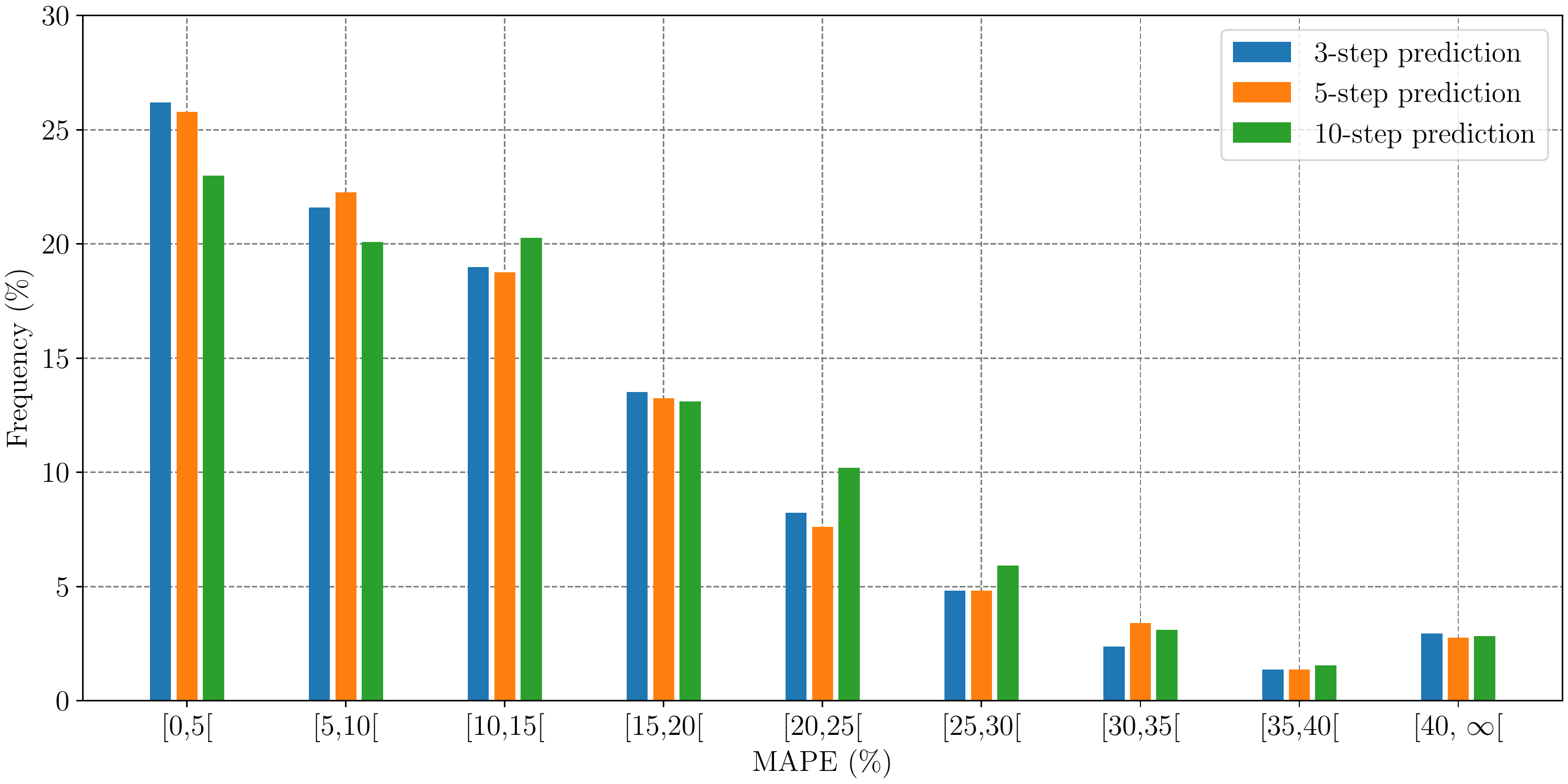}
		\vspace{-1.5em}
		\caption{Distribution of MAPE values on hidden interfaces.}\label{fig:dist_err_pred_compar_unobs}
	\end{subfigure}%
		\caption{MAPE of TRM predictions made on the test dataset, on observed and hidden road interfaces.}\label{tab:res_unobs}
	\end{figure}

We then take a closer look at how these prediction MAPE distribute themselves along the road. To do so, we compute, on each road interface, the MAPE between the TRM predictions and the measurements, and display these errors in \Cref{fig:err_pred_space}. We note that the errors are stable across the interfaces, and that therefore there is no major difference in the quality of the predictions across space. In particular, the errors obtained on the two hidden interfaces (namely, the interfaces $4$ and $7$) are similar to the errors obtained on the surrounding interfaces that were used during training. This points to the conclusion that the algorithm seems stable with respect to missing data, and therefore that it is capable of completing and yielding trustworthy predictions on missing interfaces. Note also that the errors are slightly higher in the middle of the road. This is expected since this is where the model is most constrained by the TRM dynamics (which propagates free space from right to left and occupied space from left to right) and the measurements, and therefore has to balance them the best it can.

\begin{figure}
	\centering
	\begin{subfigure}{0.33\textwidth}
		\includegraphics[width=\textwidth]{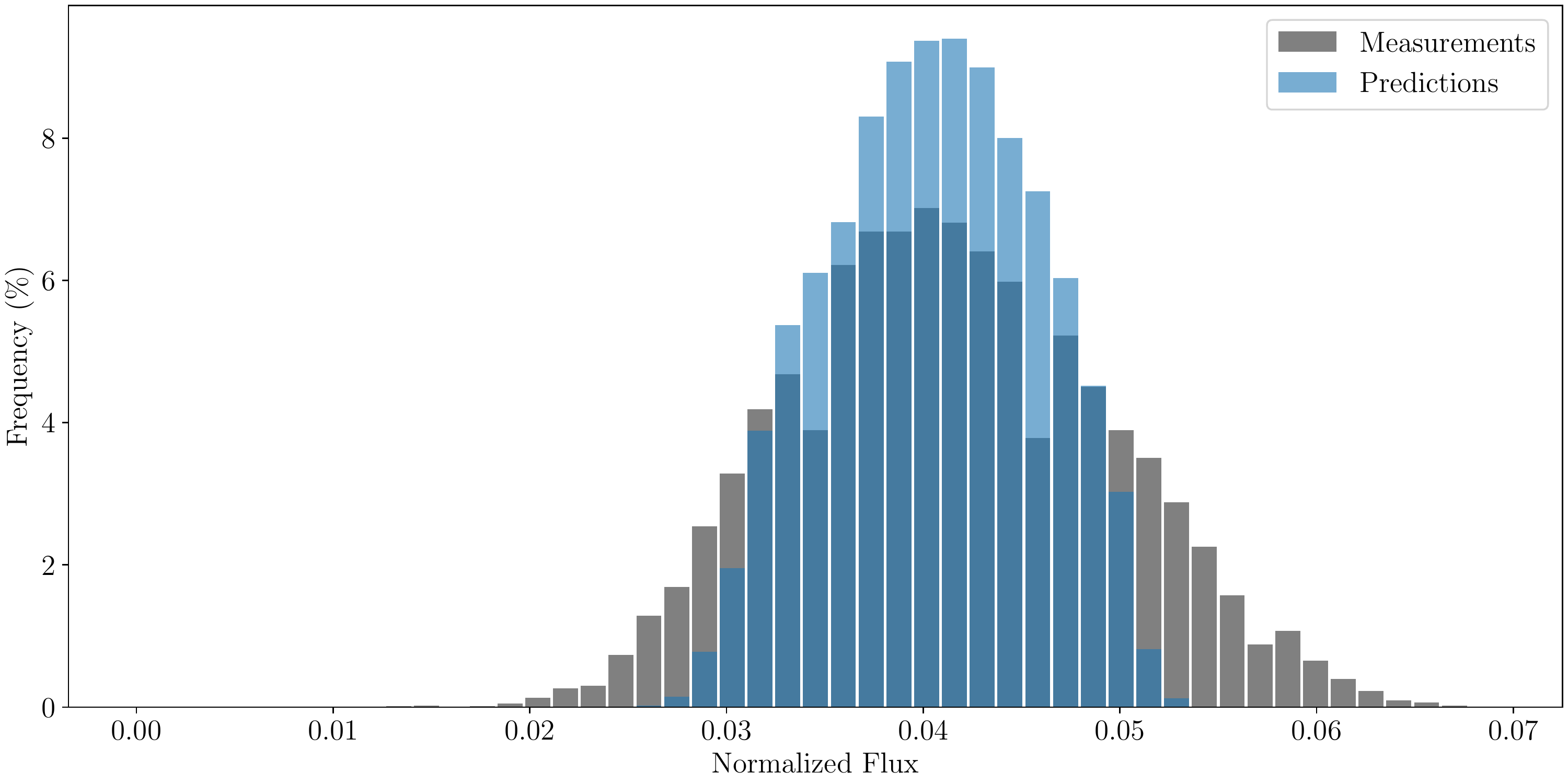}
		\vspace{-1.5em}
		\caption{Histogram comparison for \psa-step prediction.}
	\end{subfigure}%
	\hfill
	\begin{subfigure}{0.33\textwidth}
		\includegraphics[width=\textwidth]{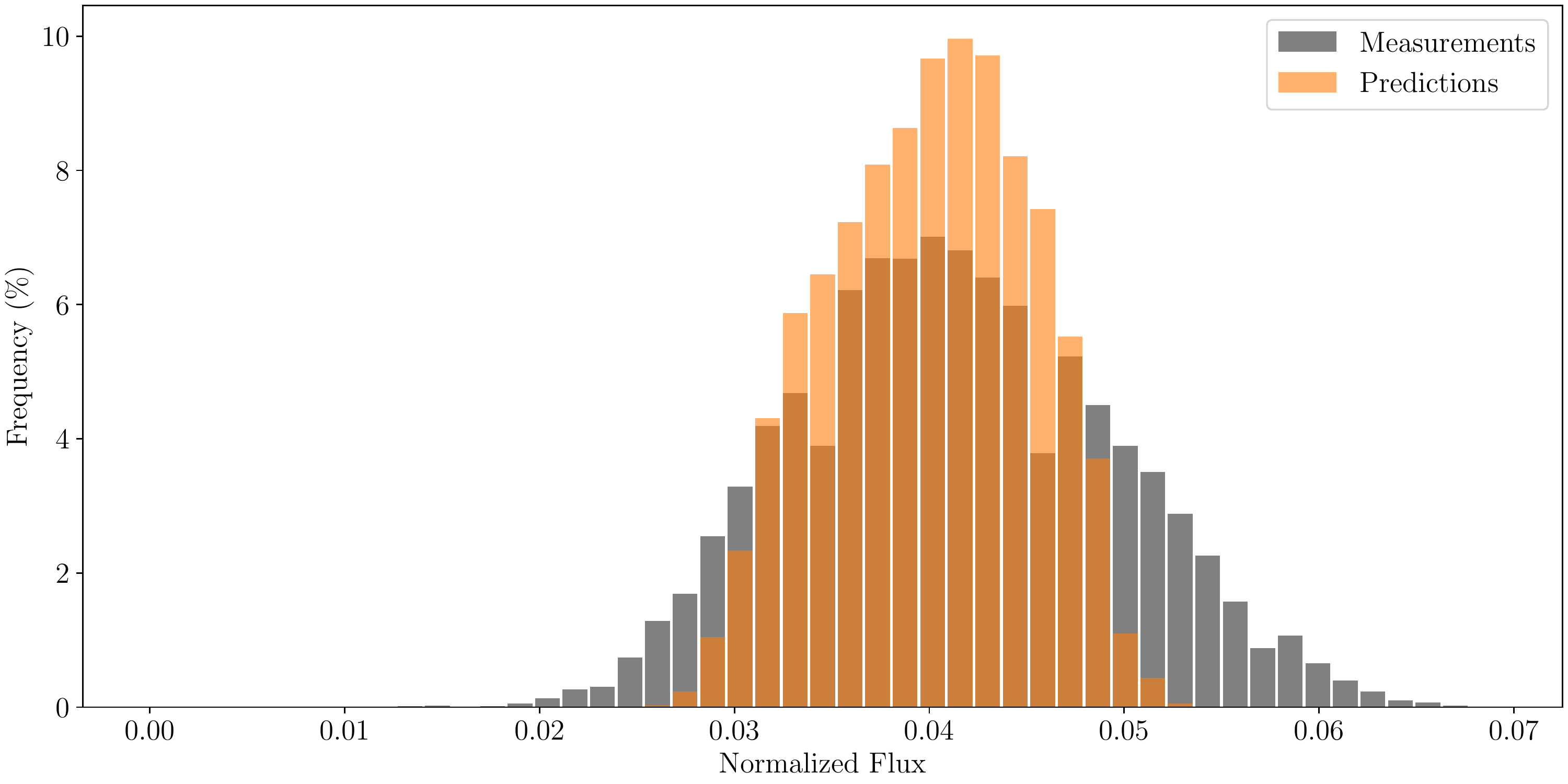}
		\vspace{-1.5em}
		\caption{Histogram comparison for \psb-step prediction.}
	\end{subfigure}%
	\hfill
	\begin{subfigure}{0.33\textwidth}
		\includegraphics[width=\textwidth]{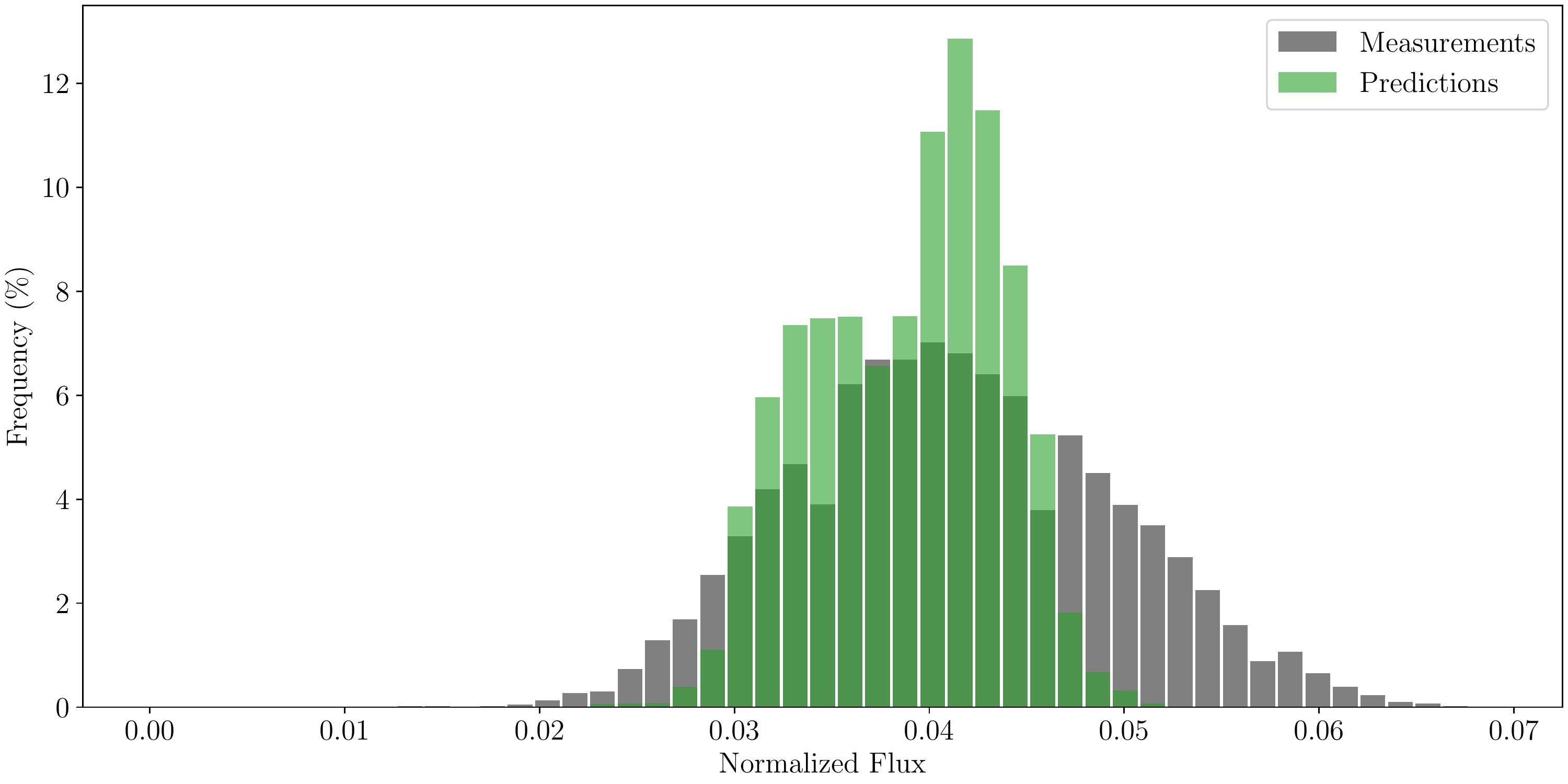}
		\vspace{-1.5em}
		\caption{Histogram comparison for \psc-step prediction.}
	\end{subfigure}
	\caption{Histogram comparison between all the available measurements (i.e.\ on both observed and hidden interfaces) and predictions (i.e.\ on all $55$ interfaces composing the road) of the test dataset.}\label{fig:hist_compar_pred_meas}
\end{figure}

\begin{figure}
	\centering
	\begin{subfigure}{0.33\textwidth}
		\includegraphics[width=\textwidth]{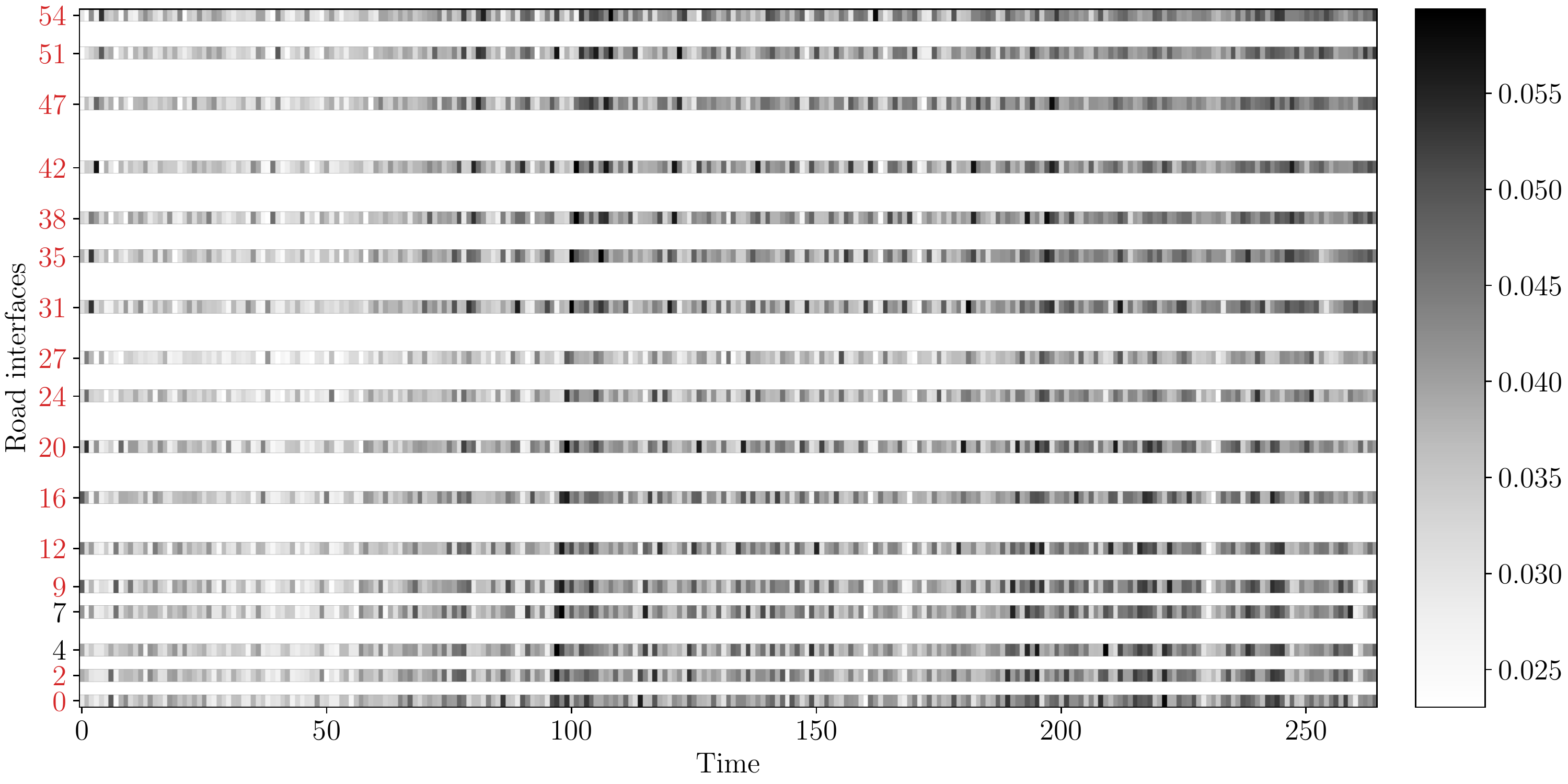}
		\vspace{-1.5em}
		\caption{Flux measurements.}
	\end{subfigure} \\
	\vspace{1em}
	\begin{subfigure}{0.33\textwidth}
		\includegraphics[width=\textwidth]{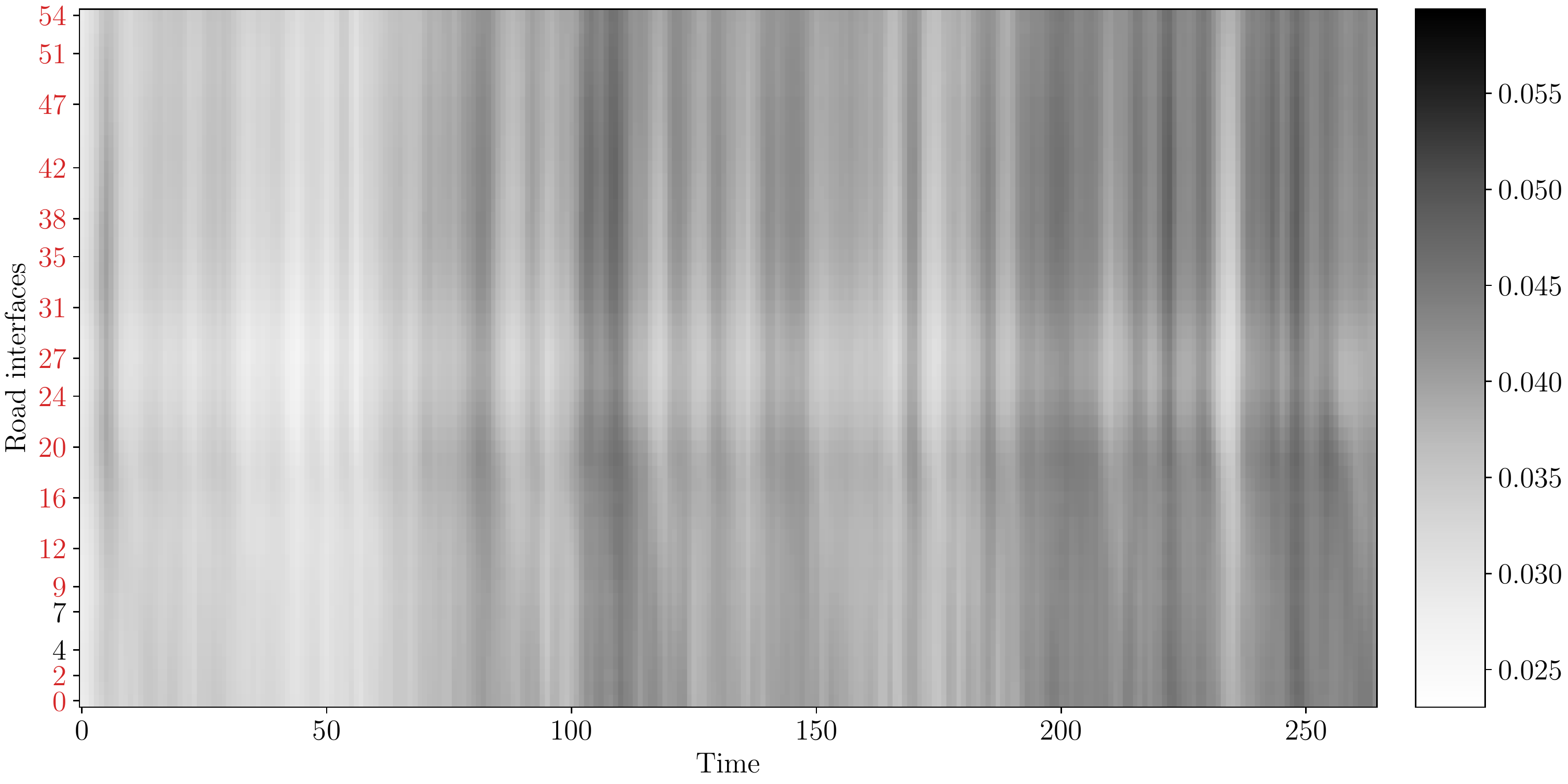}
		\vspace{-1.5em}
		\caption{\psa-step prediction.}
	\end{subfigure}%
	\hfill
	\begin{subfigure}{0.33\textwidth}
		\includegraphics[width=\textwidth]{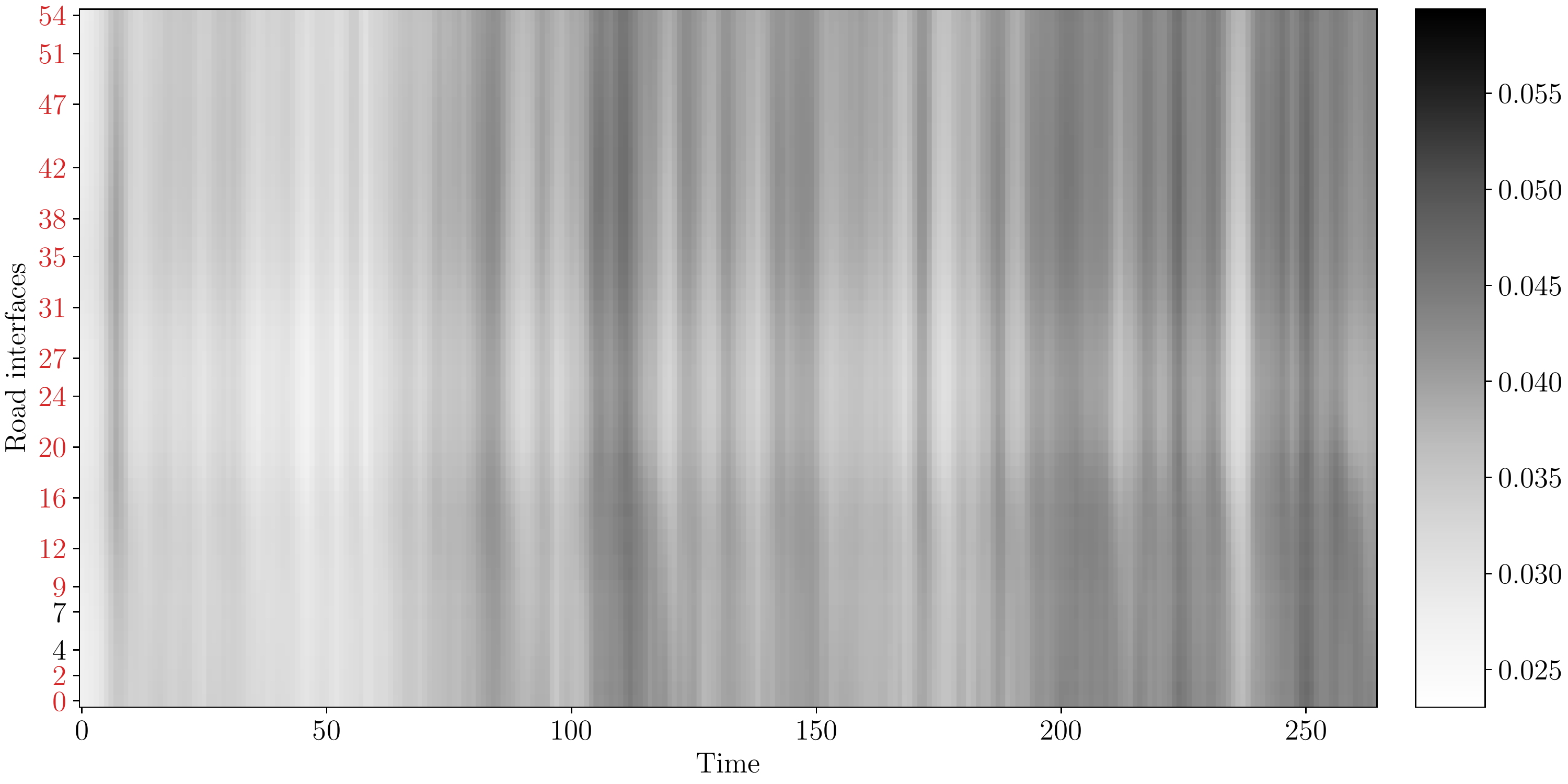}
		\vspace{-1.5em}
		\caption{\psb-step prediction.}
	\end{subfigure}%
	\hfill
	\begin{subfigure}{0.33\textwidth}
		\includegraphics[width=\textwidth]{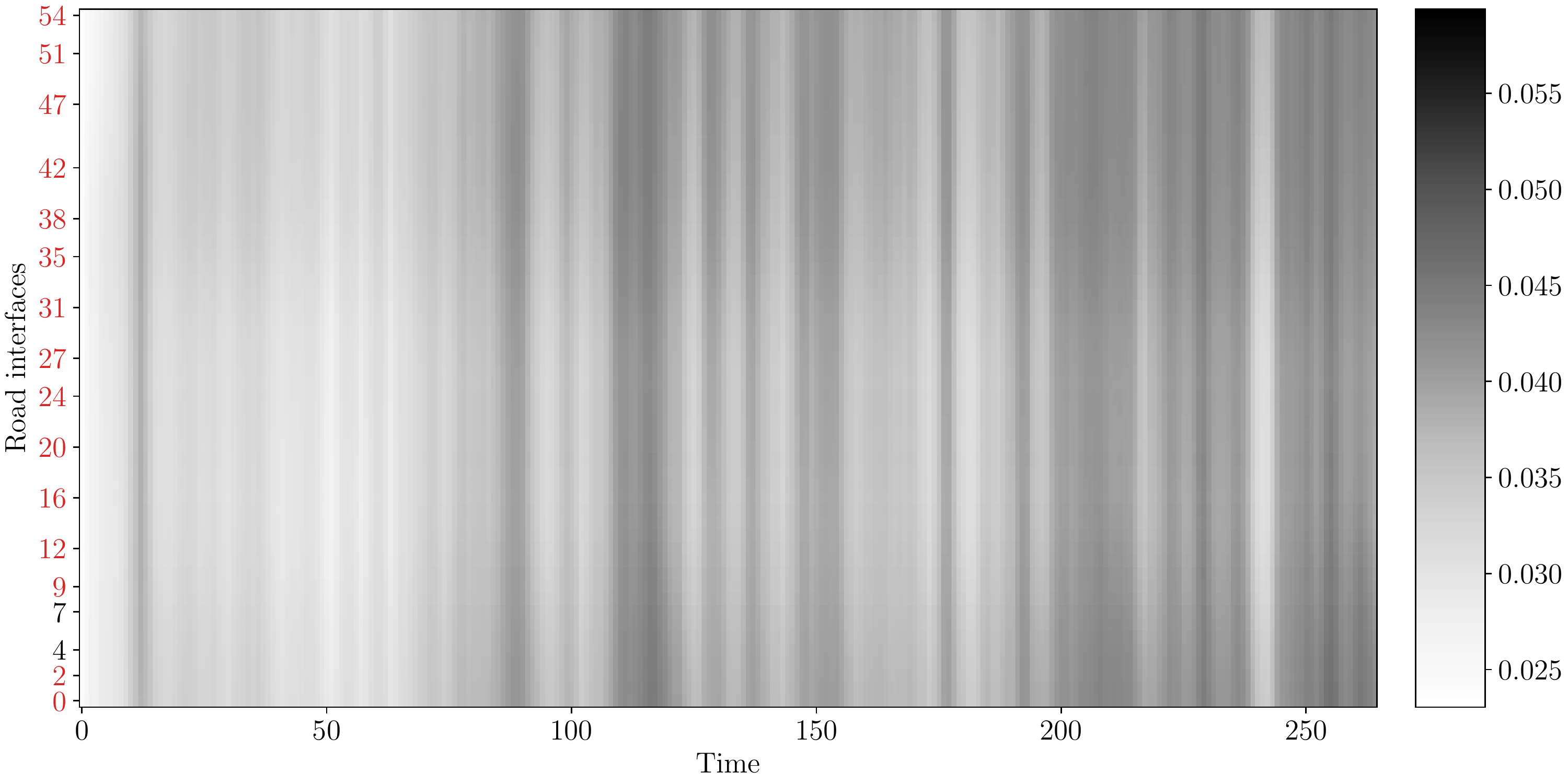}
		\vspace{-1.5em}
		\caption{\psc-step prediction.}
	\end{subfigure}%
	\caption{Comparison between the space-time variations of the measurements and predictions on May 10th. The indices marked in red correspond to the observed interfaces, and those in black to the hidden interfaces.}\label{fig:st_var_may10}
\end{figure}

We also give in \Cref{tab:err_pred_compar_obs_unobs,fig:dist_err_pred_compar_obs,fig:dist_err_pred_compar_unobs} the overall MAPE between predictions and measurements obtained on observed and hidden interfaces, and the distribution of these errors. In both cases, and for all three prediction horizons, $75\%$ to $80\%$ of the errors are below $20\%$. The remaining \q{extreme} errors could be explained by the discrepancy between the TRM that yields flux predictions based on relatively smooth traffic dynamics, as modeled by the macroscopic traffic model, and the high level of noise of the data. This fact is already noticeable on the prediction plots given in \Cref{fig:ex_pred_1,fig:ex_pred_3,fig:ex_pred_2,fig:ex_pred_4}, but also when plotting  space-time plots of measurements and predictions as in \Cref{fig:st_var_may10}. We confirm by representing in \Cref{fig:hist_compar_pred_meas} histograms comparing the distributions of measurements and predictions obtained on the test dataset: it is clear that the fat tails of the measurement distribution no longer appear on the prediction distributions. 

\begin{figure}
		\centering
		\begin{subfigure}[t]{0.4\textwidth}
			\includegraphics[width=\textwidth]{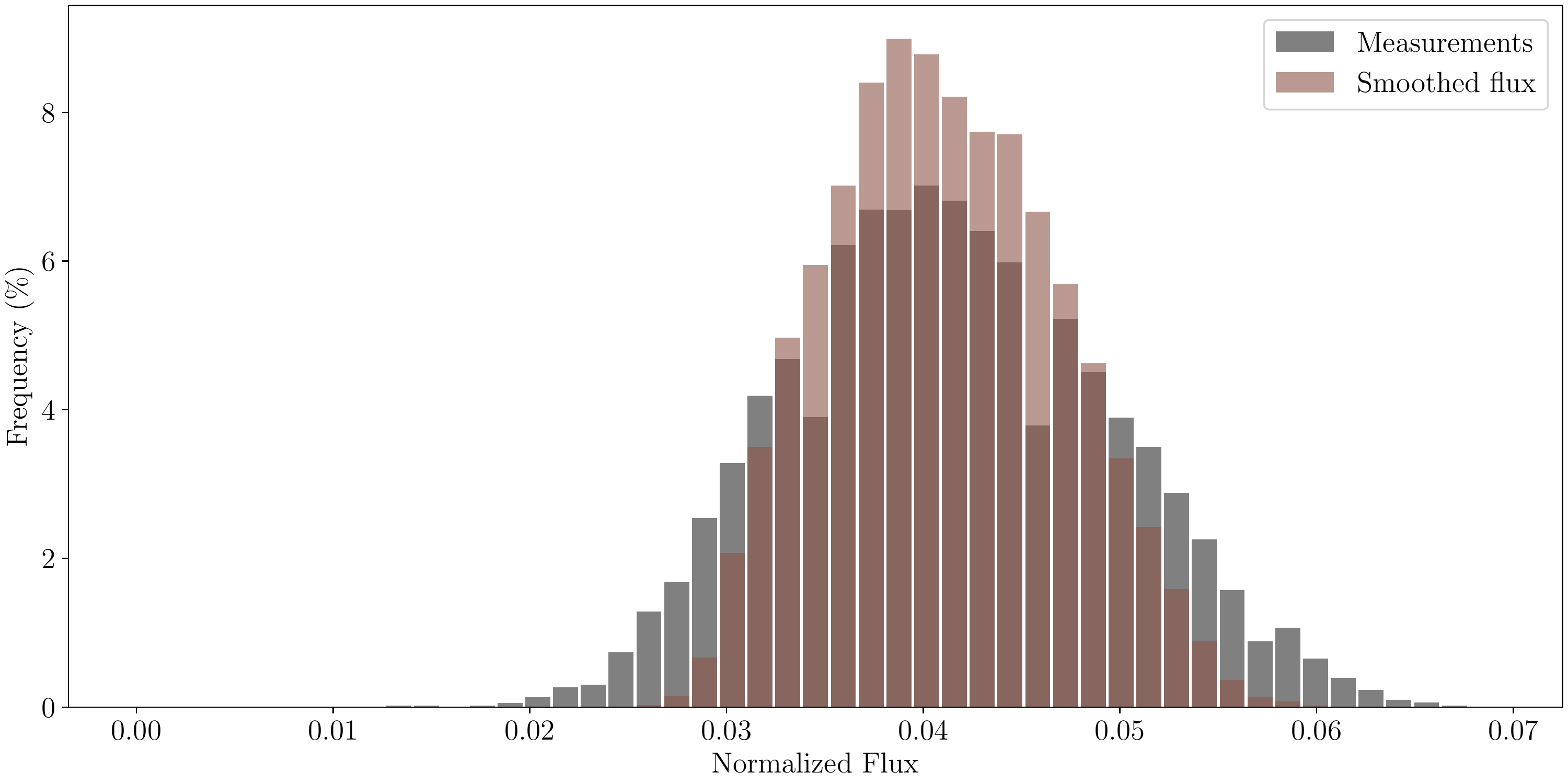}
			\vspace{-1.5em}
			\caption{Histograms of measurements and smoothed fluxes.}\label{fig:hist_compar_smooth}
		\end{subfigure}%
		\hspace{3em}
		\begin{subfigure}[t]{0.4\textwidth}
			\includegraphics[width=\textwidth]{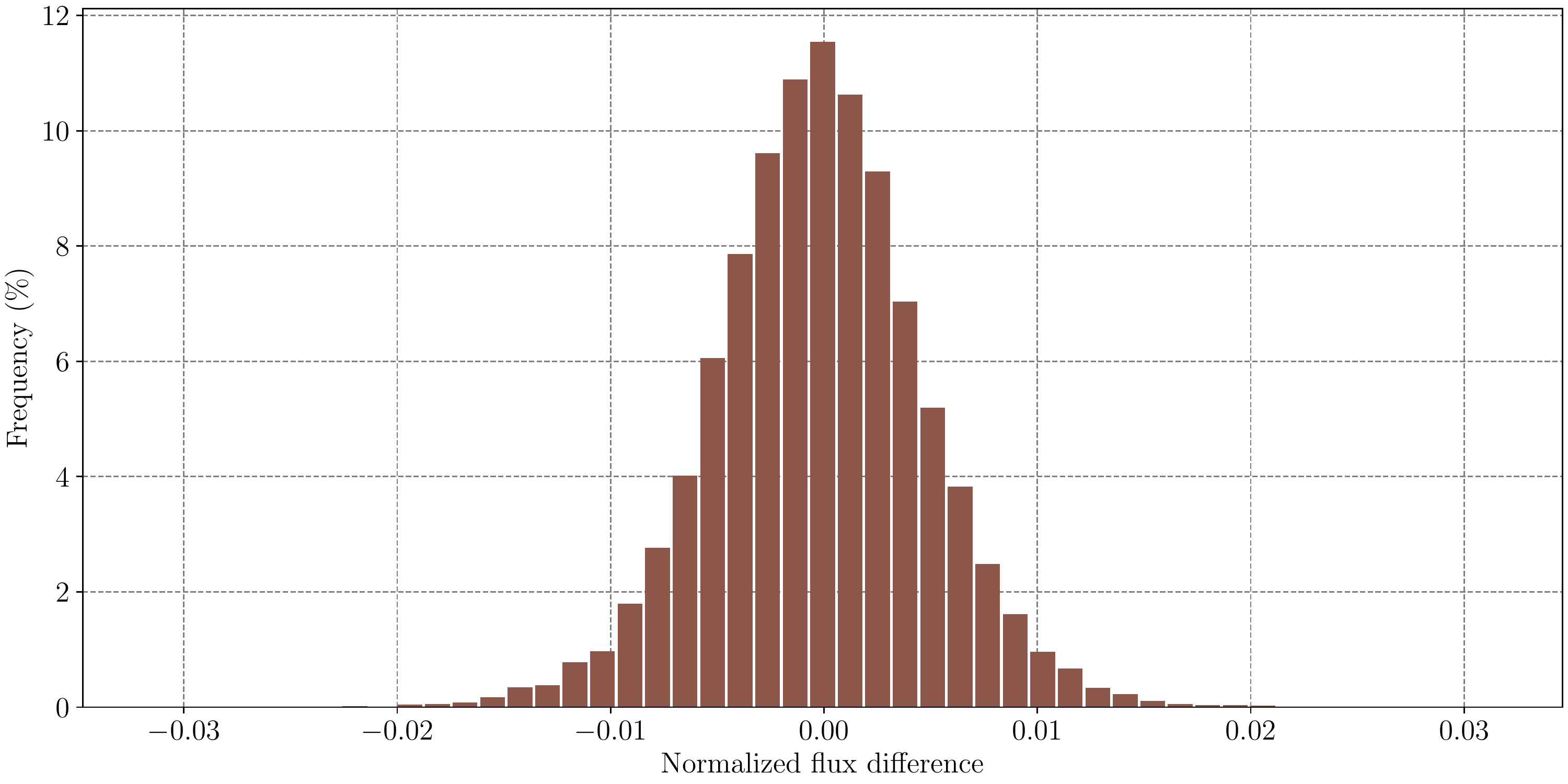}
			\vspace{-1.5em}
			\caption{Histogram of the difference between the smoothed fluxes and the measurements.}\label{fig:hist_diff_smooth}
		\end{subfigure}\\
		\begin{subfigure}[t]{0.6\textwidth}
			\centering
			\includegraphics[width=0.685\textwidth]{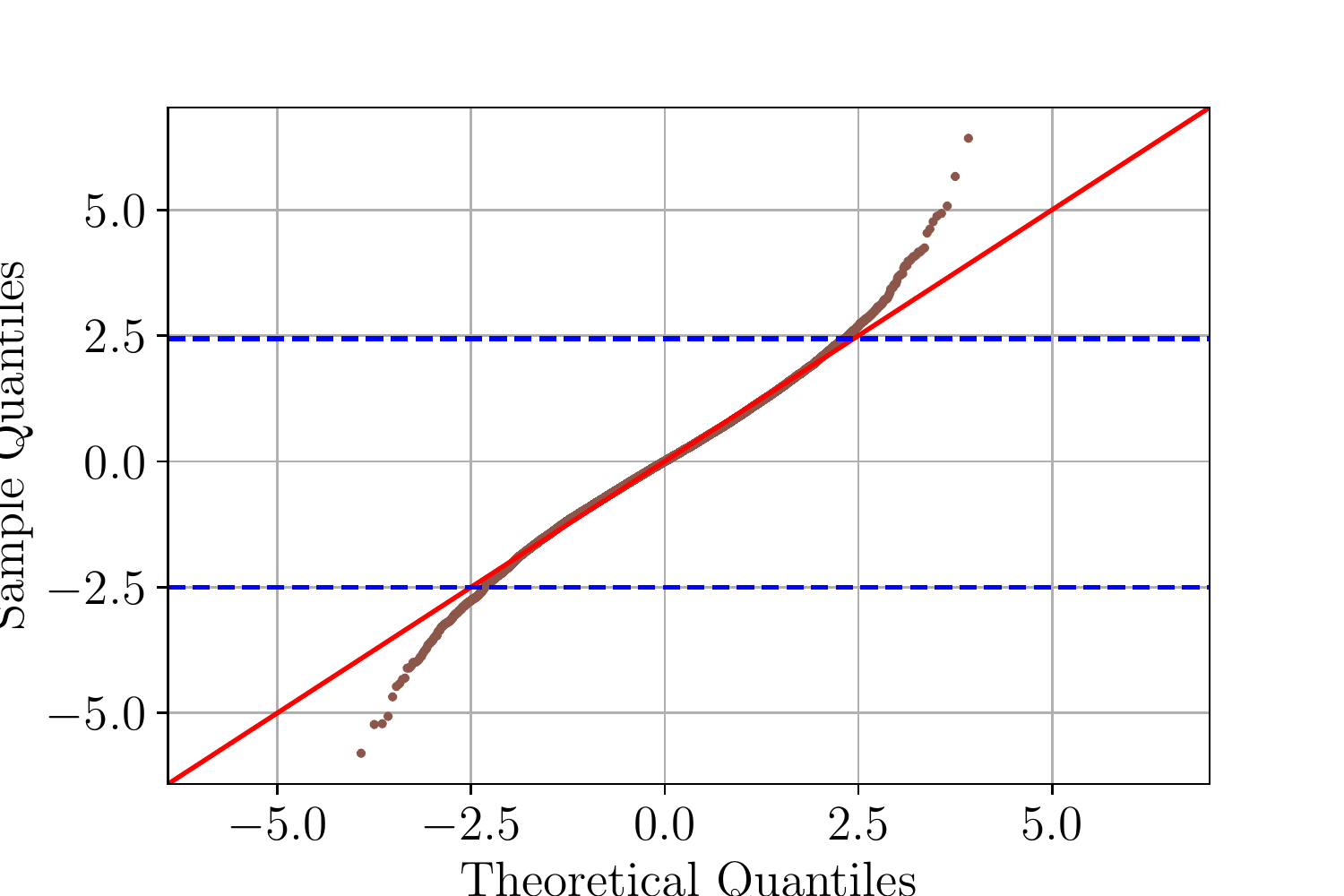}
			\caption{QQ-plot of the difference between the smoothed fluxes and the measurements (adjusted in mean in variance) with respect to the standard normal distribution. 98\% of the points lie between the two blue lines.}\label{fig:qqplot_smooth}
		\end{subfigure}
		\caption{Comparison between the measurements and the smoothed fluxes.}\label{fig:compar_smooth_meas}
	\end{figure}

	This raises the question of the level of smoothing introduced by our approach. Recall indeed that our algorithm yields predictions by recreating, under a macroscopic traffic model, the traffic dynamics as observed in the past observations (at times $t_{-14}, \dots, t_0$) and extending them to future times ($t_1, \dots, t_{10}$). In particular, at $t_0$, a smoothed version of the most recent observation is available: we refer to it as smoothed flux. This smoothed flux is for instance represented in some particular cases in \Cref{fig:ex_pred_1,fig:ex_pred_3,fig:ex_pred_2,fig:ex_pred_4}, and represents the current traffic state as seen by the macroscopic traffic model. We perform in \Cref{fig:compar_smooth_meas} a comparison between the measurements and these smoothed fluxes.
	
	First, note in the histograms in \Cref{fig:hist_compar_smooth} that, as expected, smoothed fluxes present lighter tails than the measurements. And besides, as seen in \Cref{fig:hist_diff_smooth}, when taking the difference between the smoothed fluxes and the measurements, we obtain errors with a symmetric distribution around zero: hence the smoothing does not tend to overestimate or underestimate the fluxes. When comparing the distribution of these errors to a normal distribution (cf.\ QQ-plot in \Cref{fig:qqplot_smooth}), we can see that most of the errors (98\% of them) can be compared to Gaussian errors, but the remaining ones tend to take more extreme values than Gaussian errors would. Hence, we have a non-Gaussian distribution of smoothing errors, with fatter tails than a Gaussian distribution. These fat tails reflect the ability of the algorithm to propose smoothed fluxes that are robust to outliers in the measurements.
	
\begin{figure}
	\centering
	\begin{subfigure}{0.6\textwidth}
	\centering
	\adjustbox{width=\textwidth}{
		\begin{tabular}{*{4}{C{3cm}}}
\toprule
{} &  On observed interfaces  &  On hidden interfaces &    All \\ \midrule
3-step prediction  &                   5.94\% &                4.39\% & 4.82\% \\ \hline
5-step prediction  &                   7.11\% &                6.03\% & 6.32\% \\ \hline
10-step prediction &                   8.63\% &                7.79\% & 8.02\% \\ \bottomrule
\end{tabular}

	}
	\vspace{1ex}
	\caption{MAPE between the predicted and smoothed fluxes on the observed and hidden interfaces.}\label{tab:err_pred_compar_smooth}
\end{subfigure}	  \\
\vspace{1em}
	\begin{subfigure}[t]{0.9\textwidth}
	\includegraphics[width=\textwidth]{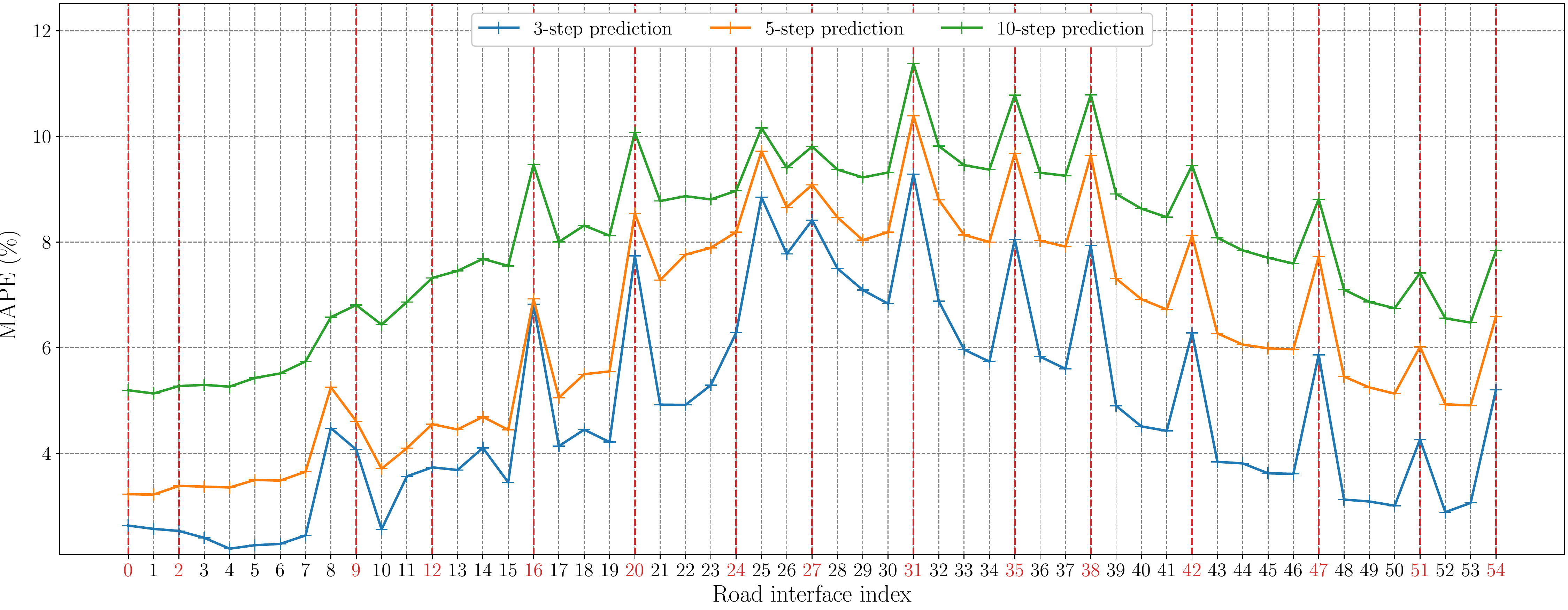}
	\vspace{-1.5em}
	\caption{MAPE on each road interface between the predicted and smoothed fluxes. The red lines mark the observed interfaces. The indices marked in red also correspond to the observed interfaces, and those in black to the hidden interfaces.}\label{fig:err_pred_smooth_space}
\end{subfigure}
	\caption{Comparison between the predictions and the smoothed fluxes computed on the test dataset.}
\end{figure}

	In conclusion, our algorithm yields smoothed fluxes that are physics-aware (since they arise from the macroscopic traffic model) and well-behaved (since they are robust to outliers and unbiased). Besides, these smoothed fluxes are available on all of the interfaces dividing the road, and not only at the locations where measurements were taken. If we now use these smoothed fluxes as ground truth when computing prediction MAPEs (instead of using the raw flux measurements), we note that we get errors that are drastically reduced compared to the case where we compare our predictions to the raw measurements (cf.\ \Cref{tab:err_pred_compar_smooth}): indeed, reductions ranging from $45\%$ to $55\%$ can be observed on the considered prediction horizons (when considering in both cases only the observed interfaces). When looking at the spatial distribution of these errors (cf.\ \Cref{fig:err_pred_smooth_space}), we retrieve the fact that they are higher further away from the boundaries, just like the prediction errors. Besides, the errors peak at the interfaces where data are observed, which can be seen as a sign that the algorithm modifies locally the dynamics to match the data recorded there (where only the dynamics of the traffic model plays a role on the other interfaces).  These observations are coherent with the fact that our algorithm was optimized to extend to the future the dynamics that gave rise to the smoothed flux: hence, the predictions obtained by our algorithm can be seen as accurate predictions of a physics-aware smoothing of the raw measurements.

\subsection{Comparison to existing methods}\label{sec:compar_ex}

Comparing the prediction performances obtained in the numerical experiment with other works is no easy task. Indeed, for the comparison to be fair, our algorithm should be compared to approaches that take as input the same type of information, namely only past raw flux measurements taken at a relatively high frequency (every minute in our case). Besides, the question of the metric used to compare different studies is not straightforward. Indeed the prediction errors are compared based on a comparison between the prediction and what is assumed to be the ground truth. However, most of the studies apply some preprocessing on the raw data (e.g.\ aggregation or smoothing) and use the result as ground truth. Since these steps usually differ from one study to the other, and are not present at all in ours, comparing the performances becomes complicated. Finally, the question of the amount of parameterization in a model and the amount of data used to train it play a central role in the performance of an algorithm, and are hard to compare from one approach to the other.

Let us omit the above comments and compare anyway the prediction performances of our algorithm with other deep neural network approaches. As noted in the review article \cite{do2019survey} , such approaches were able to yield predictions MAPE ranging from $5\%$ to $10\%$. Compared to the MAPEs we computed from the raw data (cf.\ \Cref{tab:err_pred_obs}), our algorithm seems to yield higher errors. However, compared to the MAPEs we computed from the smoothed fluxes (cf.\ \Cref{tab:err_pred_compar_smooth}) we retrieve similar values.

\section{Conclusion}

In this work, we propose a physics-aware algorithm for short-term flux predictions based a macroscropic traffic flow model. This algorithm has two main components. The first one is a RNN which aims at extracting the space-time varying parameters of the traffic flow model associated with a set of past observations and extending them to future time steps. These parameters are then passed to the second component, which is a TRM discretization of the traffic flow model: we then obtain flux estimates at past and future time steps that arise directly from the traffic flow model. A numerical experiment, conducted on a dataset of raw flux measurements taken along a stretch of freeway in the Netherlands, highlights the advantages of our approach.

First, the algorithm outputs a reestimation of the most recent measurement it got as input, which can be used as a smooth estimate for the traffic state. This is particularly desirable when dealing with raw data, for which choosing and justifying the right aggregation or smoother to reduce the noise can be cumbersome. In our case, the noise reduction is streamlined with the prediction and is physically grounded by the macroscopic traffic flow model. As for the predictions, they arise from the same traffic model and are therefore a simple continuation in time of the dynamics inferred from the past observations. Hence both the smoothing of raw data and the subsequent predictions have a clear physical interpretation. Finally, the algorithm yields smooth flux estimates and predictions everywhere on the road (and not only at the locations where measurements are made). Hence our algorithm also serves as a data augmentation tool. Future work include th extension of the approach to networks, which can be made possible by extending the TRM part of the architecture to networks.

%% References
\bibliographystyle{apalike}
\bibliography{biblio}

\appendix
	
	\section{Multilayer perceptrons}\label{sec:MLP}
	
	We provide in this section a short definition of the notions of single layer and multilayer perceptrons, and refer the interested reader to \citep{bishop2006pattern} for more details.
	
	A (single layer) perceptron is an artificial neural network (ANN) which takes as input a vector $\bm x \in\R^{\Nin}$ ($\Nin\in\mathbb{N}$) and returns an output vector $\bm y\in\R^{\Nout}$ ($\Nout \in\mathbb{N}$) whose entries $y_1, \dots, y_{\Nout}$ are given by:
	\begin{equation*}
		y_i = \sigma(\bm W^{(i)}\bm x + b_i), \quad i \in\lbrace1, \dots, \Nout\rbrace,
	\end{equation*}
	where $\bm W^{(i)}\in\R^{\Nin\times \Nout}$ is a weight matrix, $b_i\in\R$ is a bias term and $\sigma$ is a nonlinear function called activation function. Examples of activation functions commonly used in practice include the Heaviside function, the sigmoid function, the hyperbolic tangent and the ReLu function. Hence each entry of the output of a perceptron is nothing but a linear combination of the entries of its input vector to which we add a bias term and apply an activation function.
	
\begin{figure}[h]
		\centering
		\adjustbox{width=\textwidth}{
			\begin{Huge}
				\begin{tikzpicture}

\matrix[column sep=2cm, row sep=0.5cm,minimum size=2cm] (m) {
	
	\node[circle,draw] (in)  {$\bm x$};

	& \node[rectangle,draw] (s1)  {$\mathcal{S}_1$};
	
	& \node[circle,draw] (h1)  {$\bm h^{(1)}$};
	
	& \node[rectangle,draw] (s2)  {$\mathcal{S}_2$};
	
		& \node[circle,draw] (h2)  {$\bm h^{(2)}$};
	
	& \node[rectangle] (sd)  {$\cdots$};
	
		& \node[circle,draw] (hnm1)  {$\bm h^{(n-1)}$};
	
	& \node[rectangle,draw] (sn)  {$\mathcal{S}_n$};
	
	& 	\node[circle,draw] (out)  {$\bm y$};\\
};

\draw[-latex,line width=2.5] (in) -- (s1);

\draw[-latex,line width=2.5] (s1) -- (h1);

\draw[-latex,line width=2.5] (h1) -- (s2);

\draw[-latex,line width=2.5] (s2) -- (h2);

\draw[-latex,line width=2.5] (h2) -- (sd);

\draw[-latex,line width=2.5] (sd) -- (hnm1);

\draw[-latex,line width=2.5] (hnm1) -- (sn);

\draw[-latex,line width=2.5] (sn) -- (out);

\end{tikzpicture}
			\end{Huge}
		}
		\caption{Multilayer perceptron: $\mathcal{S}_1, \dots, \mathcal{S}_n$ denote single layer perceptrons, $\bm h^{(1)}, \dots, \bm h^{(n-1)}$ denote hidden layers, and circles are used to denote input and/or output vectors.}
		\label{fig:mlp}
	\end{figure}
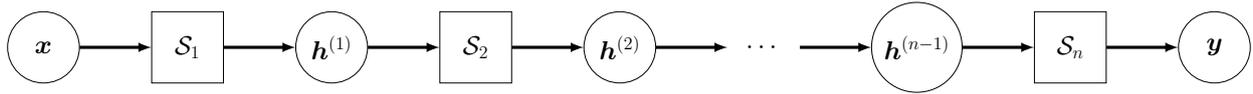

	A MultiLayer Perceptron (MLP), also called feed-forward neural
	network, is the ANN obtained by stacking together multiple (single layer) perceptrons: the output of each perceptron is used as input of the next one (cf.\ \Cref{fig:mlp}). In such a configuration, the layers between the input vector and the final output vector are called hidden layers, and single layer perceptrons can be seen as a particular case of MLP with no hidden layer.
	
\end{document}